\documentclass{article}

\PassOptionsToPackage{numbers, sort&compress}{natbib}

\usepackage[final]{neurips_2025_ml4ps}

\usepackage[utf8]{inputenc} % allow utf-8 input
\usepackage[T1]{fontenc}    % use 8-bit T1 fonts
\usepackage[colorlinks=false,hypertexnames=false]{hyperref} % hyperlinks
\usepackage{url}            % simple URL typesetting
\usepackage{booktabs}       % professional-quality tables
\usepackage{amsfonts}       % blackboard math symbols
\usepackage{nicefrac}       % compact symbols for 1/2, etc.
\usepackage{microtype}      % microtypography
\usepackage{xcolor}         % colors
 \usepackage{graphicx}      % figures
\usepackage{enumitem}       % lists
\usepackage{listings}       % code/prompt/instruction listings (no shell-escape needed)
\usepackage{standalone} % auto-crops
\usepackage{amsmath}
\usepackage{float}
\usepackage{tcolorbox}
\usepackage{textcomp}
\usepackage{cleveref}
\tcbuselibrary{listingsutf8}
\tcbuselibrary{skins,breakable}
\usepackage{array}
\newcolumntype{C}[1]{>{\centering\arraybackslash}m{#1}}

\usepackage{endnotes}

\definecolor{codeborder}{RGB}{50,100,200}      % frame
\definecolor{codebackground}{RGB}{245,245,245} % background
\definecolor{titlebg}{RGB}{232,238,247}        % title band

\tcbuselibrary{skins,breakable}
\definecolor{rwsframe}{gray}{0.30}   % frame
\definecolor{rwsback}{gray}{0.98}    % box background
\definecolor{rwstitle}{gray}{0.15}   % title band
\ifdefined\bwtrue
  \definecolor{rwsframe}{gray}{0.30}
  \definecolor{rwsback}{gray}{0.98}
  \definecolor{rwstitle}{gray}{0.15}
\fi
\tcbset{
  rwscommon/.style={
    enhanced, breakable,
    colback=rwsback, colframe=rwsframe, boxrule=0.4pt, arc=2pt,
    left=1.2mm, right=1.2mm, top=1mm, bottom=1mm,
    before skip=0.75\baselineskip, after skip=0.75\baselineskip,
    attach boxed title to top left={yshift=-2pt, yshifttext=-2pt},
    varwidth boxed title, % wrap long titles nicely
    boxed title style={colback=rwstitle, colframe=rwstitle, boxrule=0pt, arc=2pt,
      left=1.2mm, right=1.2mm, top=0.6mm, bottom=0.6mm},
    fonttitle=\bfseries\small, coltitle=white,
  }}

\definecolor{apframe}{gray}{0.28}
\definecolor{apback}{gray}{0.98}
\definecolor{aptitle}{gray}{0.15}
\definecolor{apaccent}{gray}{0.20}

\newlist{agentsteps}{enumerate}{1}
\setlist[agentsteps]{label=\textbf{\arabic*.}, leftmargin=*, itemsep=0.25em, topsep=0.25em}

\tcbset{
  agentcommon/.style={
    enhanced, breakable,
    colback=apback, colframe=apframe, boxrule=0.4pt, arc=2pt,
    left=1.6mm, right=1.6mm, top=1.2mm, bottom=1.0mm,
    borderline west={2pt}{0pt}{apaccent},
    before skip=0.75\baselineskip, after skip=0.75\baselineskip,
    attach boxed title to top left={yshift=-2pt, yshifttext=-2pt},
    varwidth boxed title, % wrap long titles cleanly
    boxed title style={
      colback=aptitle, colframe=aptitle, boxrule=0pt, arc=2pt,
      left=1.6mm, right=1.6mm, top=0.6mm, bottom=0.6mm
    },
    fonttitle=\bfseries\small, coltitle=white,
    fontupper=\small,
    before upper=\setlength{\parskip}{3pt}\setlength{\parindent}{0pt}\obeyspaces,
  }
}

\newcommand{\promptspace}{\hspace*{\fontdimen2\font}}

\newtcolorbox{agentinstruction}[1][]{agentcommon, #1}

\newtcolorbox[auto counter, number within=section]{codebox}[1][]{%
  rwscommon,
  listing only,
  listing options={
    language=Python,
    basicstyle=\ttfamily\footnotesize,
    keywordstyle=\color{black},
    commentstyle=\color{gray},
    stringstyle=\color{black},
    showstringspaces=false,
    columns=fixed,
    breaklines=true,
    inputencoding=utf8,
    upquote=true,
    literate=
      *{^}{{\char`^}}1
       {|}{{\char`|}}1
       {<}{{\char`<}}1
       {>}{{\char`>}}1
       {\{}{{\char`\{}}1
       {\}}{{\char`\}}}1
       {~}{{\char`~}}1
  }
  ,#1
}

\newtcolorbox{rwsbox}[2][]{rwscommon, title={#2}, #1}

\newfloat{example}{htbp}{loe}[section]
\floatname{example}{Example}

\title{Reasoning With a Star: A Heliophysics Dataset and Benchmark for Agentic Scientific Reasoning}

\hypersetup{pdfborder={0 0 1}}
\usepackage{orcidlink}

\author{
Kevin Lee\orcidlink{0009-0004-0388-9260}$^{1,2}$\thanks{Corresponding author. Email: \texttt{kevinlee69720@g.ucla.edu}} \quad
Russell Spiewak\orcidlink{0000-0003-0196-5712}$^{1,3}$ \quad
James Walsh\orcidlink{0000-0002-2100-7696}$^{1,3,4}$ \\
$^1$Frontier Development Lab, Frisco, TX, USA \\
$^2$Department of Mechanical and Aerospace Engineering, UCLA, Los Angeles, CA, USA \\
$^3$Trillium Technologies Inc., Frisco, TX, USA \\
$^4$Department of Engineering, University of Cambridge, Cambridge, UK
}

\begin{document}

\begingroup
  \hypersetup{hidelinks, pdfborder={0 0 0}}
  \renewcommand{\thefootnote}{\fnsymbol{footnote}}
  \maketitle
\endgroup

\begin{abstract}
Scientific reasoning through Large Language Models in heliophysics involves more than just recalling facts: it requires incorporating physical assumptions, maintaining consistent units, and providing clear scientific formats through coordinated approaches. To address these challenges, we present Reasoning With a Star, a newly contributed heliophysics dataset applicable to reasoning; we also provide an initial benchmarking approach. Our data are constructed from National Aeronautics and Space Administration \& University Corporation for Atmospheric Research Living With a Star summer school problem sets and compiled into a readily consumable question-and-answer structure with question contexts, reasoning steps, expected answer type, ground-truth targets, format hints, and metadata. A programmatic grader checks the predictions using unit-aware numerical tolerance, symbolic equivalence, and schema validation. We benchmark a single-shot baseline and four multi-agent patterns, finding that decomposing workflows through systems engineering principles outperforms direct prompting on problems requiring deductive reasoning rather than pure inductive recall.
\end{abstract}

\section{Introduction}
Scientific problem solving rarely fits into a single logical leap, especially in fields requiring deep prior knowledge. Progress usually requires domain expertise, institutional resources, iterative refinements, and validation of assumptions. This coincides with systems engineering practices, which emphasize well-defined interfaces, requirements, and verifications \cite{hirshorn_nasa_2017}. Meanwhile, Large Language Models (LLMs) exhibit algorithmic limitations in reasoning, causing ``reasoning illusions'' \cite{shojaee_illusion_2025} and algebraic failures. This motivates multi-step, role-based reasoning approaches with chain-of-thought \cite{wei_chain--thought_2023}, tree-of-thoughts \cite{yao_tree_2023}, and graph-of-thoughts \cite{besta_graph_2024}, emulating how scientists iteratively formulate hypotheses, refine models, and validate results, rather than one-shot guessing.

To bridge this gap, we present Reasoning With a Star (RWS)%
\begingroup
  \renewcommand{\thefootnote}{\fnsymbol{footnote}}%
  \footnote[2]{Available at: \href{https://huggingface.co/datasets/SpaceML/ReasoningWithAStar}{\texttt{Hugging Face - SpaceML/ReasoningWithAStar}}}%
\endgroup, a benchmark dataset for LLM- and agent-based reasoning in heliophysics derived from NASA/UCAR Living With a Star (LWS) Summer School problem sets \cite{nasa_school_2025, nasa_textbook_2025}. Drawing from systems engineering design practices in mission-critical environments, we evaluate multi-agent systems designed with manageable complexity through the Systems-engineering-of-Thoughts Agentic Reasoning (STAR, \appendixautorefname{}~\ref{appendix:star}). Using a programmatic grader for symbolic equivalence, unit-aware numerical tolerance, and schema validation, our experiments show that no single agentic pattern excels in all scenarios.

We assessed LLM- and agent-based reasoning without retrieval-augmented generation (RAG). Beyond single-shot prompting, we evaluated four agentic patterns: hierarchical multi-agent workflows (HMAW) \cite{liu_towards_2025}, plan answer critique enclose (PACE), plan hypothesize analyze solve evaluate (PHASE), and a systems-engineering-inspired expert system, informed by prior research \cite{zhao_sirius_2025,ke_mas-zero_2025} (SCHEMA). These design philosophies range from self-critique loops to assumption and requirement tracking. We find no single pattern excels in all scenarios; compact pipelines perform well on arithmetic tasks, while sophisticated patterns are better suited for methodological formulation and validation tasks.

Our contributions are threefold: (i) a science-focused benchmark; (ii) a benchmark grader adaptable to different task formats; and (iii) a comparative study of single-shot and multi-agent reasoning.

\section{Background and Related Work}

Agents are modular units that guide LLMs through specialized prompts or tools, enabling interaction with external data and the execution of complex tasks beyond text generation. When composed into workflow patterns, agents collaborate through planning, solving, verifying, or refining to support multi-step tasks beyond single-shot prompting.

A core mantra in systems engineering is that complexity must be earned, not assumed \cite{incose_complexity_primer_2015}. Excess complexity increases failure modes and reduces maintainability \cite{hirshorn_nasa_2017}. Our agentic designs follow this principle by prioritizing clear interfaces, compact workflows, and careful scaling, as demonstrated by SCHEMA’s limited use of expert agents and assumption tracking. PACE and PHASE explore the trade-offs between complexity and performance through self-critique and hypothesis formation.

We evaluated RWS alongside established reasoning benchmarks: GSM8K \cite{cobbe_training_2021}; MATH \cite{hendrycks_measuring_2021}; GPQA \cite{rein_gpqa_2023}; HumanEval \cite{chen_evaluating_2021}; and SWE-bench Verified \cite{jimenez_swe-bench_2024}. While LWS is designed to educate heliophysics researchers, RWS is well suited for training and evaluating LLMs in scientific reasoning.

\section{Dataset}
\label{sec:Data}

% RWS exemplar items (place in §Data)
\begin{figure}[H]
% \begin{example}
  \centering

  % ---------- Example A: symbolic ----------
  \begin{minipage}{0.95\linewidth}
  \begin{rwsbox}{RWS Dataset, Symbolic Instance}
  \textbf{Problem:} Interstellar gas enters the heliosphere under the influence of solar gravity, radiation pressure, and ionization losses. The resulting neutral atom density is \(n(r, \theta)\), where \(r\) is heliocentric radial distance and \(\theta\) is the angle of the heliocentric position vector relative to the bulk inflow velocity of the atoms. We may assume that the ionization rate per atom is \(\beta_{0}\left(r_{0} / r\right)^{2}\). When an atom is ionized, it has a speed approximately equal to the solar wind speed \(V\) in the frame of the solar wind. We assume that these ions are immediately picked up by the solar wind via gyration and pitch-angle scattering to form an isotropic shell of speed \(V\) in the solar wind frame.
  
  \medskip
  Assuming that the pitch-angle scattering rate is so large that the spatial diffusion tensor is negligible, write down the Parker equation for the evolution of the pickup ion omnidirectional distribution function \(f(r, \theta, v)\) with an appropriate source term. We assume that the configuration is stationary and that the solar wind has constant speed and spherical symmetry.

  \medskip
  \textbf{Solution:}
  
  \medskip
  \textbf{Step 1:} \(\frac{\partial f}{\partial t}+\mathbf{V} \cdot \nabla f-\frac{1}{3} \nabla \cdot \mathbf{V} v \frac{\partial f}{\partial v}=\beta_{0}\left(\frac{r_{0}}{r}\right)^{2} n(r, \theta) \frac{\delta(v-V)}{4 \pi v^{2}}\)

  \medskip
  \textbf{Step 2:} When integrated over \(d^{3} \mathrm{v}\), the RHS gives the rate of pickup ion generation by ionization: \(\frac{\partial n_{p i}}{\partial t}=\beta_{0}\left(\frac{r_{0}}{r}\right)^{2} n(r, \theta)\)

  \medskip
  \textbf{Step 3:} Under the specified assumptions, \(\frac{\partial f}{\partial r}-\frac{2}{3 r} v \frac{\partial f}{\partial v}=\beta_{0}\left(\frac{r_{0}}{r}\right)^{2} n(r, \theta) \frac{\delta(v-V)}{4 \pi V^{3}}\)
  
  \medskip
  \textbf{Final:} \(\boxed{\frac{\partial f}{\partial r}-\frac{2}{3 r} v \frac{\partial f}{\partial v}=\beta_{0}\left(\frac{r_{0}}{r}\right)^{2} n(r, \theta) \frac{\delta(v-V)}{4 \pi V^{3}}}\)
  \end{rwsbox}
  \end{minipage}

  \caption{Example symbolic item from the Reasoning With a Star (RWS) dataset, drawn from \texttt{2010\_Lee\_hw.pdf} \cite{nasa_textbook_2025}, showing a problem, reasoning steps, and a \LaTeX{} final expression.}
  \label{fig:rws-examples}
\end{figure}
% \end{example}

Heliophysics is the study of how the Sun affects Earth, planetary environments, and space weather, with consequences for climate, technological systems, and operational safety in space and on the ground. Despite the field's importance for satellites, communication networks, and other critical infrastructure, heliophysics remains underrepresented in reasoning benchmarks for LLMs. RWS addresses this gap by providing a benchmark for scientific reasoning rather than mere fact recall.

Using an OCR-based method, we converted LWS Summer School problem sets into a machine-readable format, manually cleaned up scanning errors, symbol and unit misreads, and typos, and then normalized the results into a JSON Lines (JSONL) dataset following the schema in \appendixautorefname{}~\ref{appendix:schema}. The resulting RWS dataset contains 158 question–answer pairs authored by heliophysicists (contributors listed in \Cref{appendix:lwsauthors}).

For each item, we store the problem statement (and, when applicable, a preamble summarizing previous sub-questions), a sequence of intermediate reasoning steps describing the expert solution, and the final answer in the \texttt{final} field, together with a \texttt{type} label, an optional \texttt{hint} specifying the expected output format (e.g., required units, JSON structure, or \LaTeX{} equation), and a \texttt{meta} container for metadata. Required physical assumptions (e.g., adiabatic expansion, constant acceleration, neglect of certain loss terms) are preserved in the question and step text. The \texttt{step} field serves as a reference reasoning process trace for analyzing model behavior and benchmark performance, but models are evaluated only on the \texttt{final} answer. The \texttt{final} field spans three types of ground truth: (i) \emph{numeric} answers, where the model must return one or more scalar values in the required physical units (38 items); (ii) \emph{symbolic} answers, where the model must produce a \LaTeX{}-formatted algebraic expression or equation (52 items); and (iii) \emph{textual} answers, where the model must provide a scientific phrase or qualitative statement (68 items). These categories reflect the range of outputs expected in heliophysics problem-solving, from quantitative estimates with units to derivations and physical explanations.

\section{Benchmark}\label{sec:benchmarking}

Our evaluation of RWS includes both single-shot baselines and multi-agent patterns. For the single-shot setup, we tested the base models Google Gemini~2.5~Pro~\cite{comanici_gemini_2025}, OpenAI OSS 20B and OSS 120B~\cite{openai2025gptoss120bgptoss20bmodel}, Meta Llama~3.3~\cite{grattafiori_llama_2024}, and Mistral~24.11~\cite{jiang_mistral_2023}; results are reported in \Cref{tab:benchmark_single}. To compare performance across different multi-agent patterns, we also evaluated their performance on established benchmarks, including GSM8K~\cite{cobbe_training_2021}, MATH~\cite{hendrycks_measuring_2021}, GPQA~\cite{rein_gpqa_2023}, HumanEval~\cite{chen_evaluating_2021}, and SWE-bench Verified~\cite{jimenez_swe-bench_2024}. These cross-benchmark results are shown in \tableautorefname{}~\ref{tab:benchmark_mas}.

\subsection{Methods}

We benchmarked four distinct agentic patterns: HMAW \cite{liu_towards_2025}, a lightweight hierarchical CEO/manager/worker handoff, PACE, which generates an answer and then performs a self-critique loop, PHASE, which adds an explicit hypothesis stage before solving, and SCHEMA, a systems engineering inspired expert allocation strategy informed by prior multi-agent architectures \cite{zhao_sirius_2025,ke_mas-zero_2025}.

Each pattern decomposes the task across role-specific agents, passes intermediate outputs between them, and produces a final answer. We also include a \emph{single-shot} baseline, in which a single LLM directly generates the final answer without coordination.

For RWS, we assess it without heliophysics-specific RAG; both single-shot and agentic pipelines must solve problems with only the provided problem statement. This setup isolates scientific reasoning ability, such as deriving relationships, propagating units, and stating assumptions, rather than rewarding access to external domain knowledge.

We score model outputs with a grader aligned with the RWS metadata. Each RWS item is labeled with an expected answer type (numeric, symbolic, or textual), required units, and formatting constraints. We evaluate: (i) \emph{numeric} answers by checking whether predicted scalar values match the ground truth within an acceptable error bound and include the required units; (ii) \emph{symbolic} answers using a computer algebra system (CAS; e.g., SymPy \cite{meurer_sympy_2017}) to verify algebraic equivalence to the LaTeX expression; and (iii) \emph{textual} answers by verifying that they provide scientifically accurate statements that satisfy required assumptions.

When the automatic grader flags a mismatch, for instance, when algebraically equivalent expressions fail a strict string match or when scientific statements are paraphrased, we apply an additional verifier built from two LLM agents running Gemini~2.5~Pro at temperature $=1.0$.

The \emph{Parser} agent extracts and normalizes the model output and the ground-truth answer (e.g., strips \texttt{\$...\$}, canonicalizes units, and produces \texttt{pred\_norm}, \texttt{gt\_norm}, and \texttt{type}).

The \emph{Judge} agent then applies a type-specific check. For numeric answers, it enforces a 5\% error tolerance and correct units. For symbolic answers, it decides whether two expressions are algebraically equivalent. Finally, for textual answers, it tests for strict semantic equivalence.

Unless otherwise noted, we report accuracy under this evaluator. Multi-agent systems and single-shot baselines are run under matched decoding conditions. No tools are used except for coding benchmarks, where we reuse the Docker-based evaluation harness (e.g., for \textit{SWE-bench Verified}) and report the resolution rate. Our multi-agent system interacts with the repository through an adapter to \textit{SWE-agent} \cite{yang_swe-agent_2024}, employing file-editing tools to construct compatible software patches.

\subsection{Results}
Gemini~2.5~Pro achieves the highest single-shot accuracy on RWS, as shown in \tableautorefname{}~\ref{tab:benchmark_single}. Among open-source models, OpenAI OSS~20B and OSS~120B perform next best, followed by Meta Llama~3.3 and Mistral~24.11.
This establishes a reasonably strong direct-prompting baseline without any multi-agent orchestration. We use this baseline to contextualize the gains from the coordination strategies.

\begin{table}[H]
  \centering
  \caption{Single-shot accuracy (\%) on RWS across models (best in bold).}
  \begin{tabular}{lc}
    \toprule
    Model & Accuracy \\
    \midrule
    Google Gemini 2.5 Pro & \textbf{35.44} \\
    OpenAI OSS 20B        & 32.91 \\
    OpenAI OSS 120B       & 32.91 \\
    Meta Llama 3.3        & 31.01 \\
    Mistral 24.11         & 27.22 \\
    \bottomrule
  \end{tabular}
  \label{tab:benchmark_single}
\end{table}

\begin{table}[H]
  \centering
  \caption{Accuracy (\%) by agent design pattern on each benchmark under Google Gemini 2.5 Pro (best per row in bold). The bottom row reports the unweighted macro-mean across datasets.}
  \begin{tabular}{lcccc}
    \toprule
    Dataset & HMAW & PACE & PHASE & SCHEMA \\
    \midrule
    GSM8K      & 91.05 & \textbf{93.41} & 92.35 & 86.36 \\
    MATH       & 78.31 & \textbf{81.51} & 77.84 & 71.39 \\
    GPQA       & \textbf{79.01} & 77.10 & 77.16 & 73.36 \\
    RWS        & 39.52 & 41.92 & 42.51 & \textbf{44.31} \\
    HumanEval  & 30.49 & 37.80 & 35.98 & \textbf{43.29} \\
    SWE-bench Verified  & 53.81 & 55.70 & 60.54 & \textbf{63.23} \\
    \midrule
    \textit{Macro-mean} & 62.03 & \textit{64.57} & \textit{64.40} & \textit{63.66} \\
    \bottomrule
  \end{tabular}
  \label{tab:benchmark_mas}
\end{table}

\tableautorefname{}~\ref{tab:benchmark_mas} summarizes accuracy by agent pattern in GSM8K, MATH, GPQA, RWS, HumanEval, and SWE-bench Verified, all under Gemini~2.5~Pro as the base model.
No single coordination pattern dominates all tasks.

\textbf{PACE} achieves the highest accuracy on GSM8K and MATH. These datasets emphasize multi-step arithmetic problem solving; PACE’s compact plan--answer--critique loop shows that a lightweight self-critique pipeline \cite{gou_critic_2024,lee_revise_2025,tan_improved_2025,shinn_reflexion_2023} is often enough to correct routine algebraic or calculation errors without adding extra coordination overhead.

\textbf{HMAW} leads in GPQA. GPQA involves graduate-level science QA, where a simple hierarchical manager/worker handoff appears sufficient to maintain focus on classification-style judgments and factual recall. This indicates that more complex hypothesis stages or detailed assumption tracking offer only marginal additional benefit for narrowly defined question-answering tasks.

\textbf{SCHEMA} performs best on HumanEval, SWE-bench Verified, and RWS. All three settings need outputs that meet specific format and constraint requirements. HumanEval demands executable code that passes reference tests, SWE-bench Verified requires task-specific bug fixes that maintain overall correctness in the repository, and RWS requires physically consistent answers with the correct units, stated assumptions, and final form. SCHEMA’s design focuses on coordination and verification, including clear requirement tracking and interface checks inspired by systems engineering, which helps prevent unnoticed changes and catches missing assumptions before returning a final answer.

\textbf{PHASE} yields competitive results across datasets but rarely achieves the best performance. Its clear hypothesis stage can reveal assumptions early, which is beneficial for scientific reasoning tasks. That said, additional steps also increase the probability of the reasoning deviating from the correct solution in narrowly defined problems.

Overall, these results reinforce the systems engineering principle that complexity must be earned, not assumed. Adding more roles or stages does not necessarily lead to higher accuracy. The effective pattern depends on the structure of the task: compact self-reflection (PACE) tends to excel on arithmetic-style reasoning; a minimal hierarchy (HMAW) works well for fact-heavy classification and science QA; and structured, interface-aware coordination (SCHEMA) is most valuable for domains like heliophysics and code-oriented benchmarks such as HumanEval and SWE-bench Verified, where task success relies on satisfying clear requirements, not just producing a plausible-looking answer. On RWS specifically, all multi-agent strategies perform better than their single-shot baselines, suggesting that even modest coordination can enhance scientific reasoning without any domain-specific RAG.

\section{Conclusion}
We presented Reasoning With a Star, a heliophysics benchmark based on NASA/UCAR Living With a Star problem sets and designed with a standardized schema and programmatic grader. The benchmark evaluates scientific reasoning in scenarios where models must state assumptions, keep units consistent, and deliver answers in the right formats. We used RWS to evaluate single-shot LLM prompting and compact multi-agent coordination patterns.

Our results show that no single coordination strategy is universally superior. For math-style reasoning tasks such as GSM8K and MATH, compact orchestration patterns like PACE, which plan, answer, and then perform a diagnostic self-critique, are usually sufficient. In contrast, for problems that require managing physical assumptions, ensuring unit consistency, or producing structured outputs, as in RWS and code-oriented benchmarks such as HumanEval and SWE-bench Verified, more sophisticated designs like SCHEMA perform better. SCHEMA’s clear role assignment, assumption tracking, and verification tracking help identify missing assumptions and prevent unnoticed issues, even without any heliophysics-specific RAG. These findings support the systems engineering design philosophy: complexity must be earned, not assumed. Adding more stages or agents does not automatically improve accuracy; effective design depends on the task’s requirements.

Taken together, this work provides (i) a domain-grounded benchmark for heliophysics reasoning, (ii) an automatic grading system that checks unit-aware numerical tolerance, symbolic equivalence, and schema validity, and (iii) a comparative evaluation of multi-agent patterns under matched conditions and without domain-specific RAG. In addition to reporting accuracy, this comparison opens up a clear opportunity to explore ways of coordinating reasoning tasks on unfamiliar scientific problems.

We anticipate that RWS would support more agent-based workflows for space science and space weather analysis, including tasks that connect solar activity to downstream effects in near-Earth and planetary environments. We also present RWS-driven development and benchmarking use cases in \appendixautorefname{}~\ref{appendix:use-case-mission-app}, showing how the benchmark can be integrated into broader LLM and agentic systems. We would expand RWS with additional heliophysics problem sets and improve the usability of the benchmark with more sophisticated output-format guidance and failure annotations, such as unit mismatches, unstated assumptions, and formatting violations, all in an effort to make LLM reasoning more auditable.

\newpage
\section*{Acknowledgments}
    This work is a research product of \href{https://heliolab.ai}{Heliolab}, an initiative of the \href{https://FDL.ai}{Frontier Development Lab}, delivered by Trillium Technologies in partnership with NASA, Google Cloud, and NVIDIA. This material is based upon work supported by NASA under award No. 80GSFC23CA040. Any opinions, findings, and conclusions or recommendations expressed are those of the author(s) and do not necessarily reflect the views of the National Aeronautics and Space Administration.

\bibliographystyle{ieeetr}
\bibliography{bib}

\appendix

\newpage
\section{Dataset}
\label{appendix:dataset}

\subsection{Dataset Schema}
\label{appendix:schema}

\begin{table}[H]
    \centering
    \caption{\textbf{JSONL schema for the RWS dataset.} Each record is a single question–and-answer pair with a machine-checkable target (\texttt{final}) and a type label that drives automatic grading: \texttt{numeric} (numeric value), \texttt{symbolic} (symbolic equivalence), or \texttt{textual} (textual equivalence).}
    \label{tab:fields}
    \begin{tabular}{ccC{0.58\textwidth}}
        \toprule
        \textbf{Field Name} & \textbf{Type} & \textbf{Description}\\
        \midrule
        \texttt{id} & String & Unique identifier for the QA set (e.g., \texttt{"2010\_Lee\_4\_a"}).\\
        \texttt{q\_id} & Integer & Question identifier for the QA set from the original problem (e.g., 1, 2, 3).\\
        \texttt{sub\_id} & String & Sub-question identifier for the QA set from the original problem (e.g., a, b, c).\\
        \texttt{preamble} & Array[String] & Optional ordered list of previous sub-QA sets. Each element is a short string (prior question and step/answer).\\
        \texttt{question} & String & The problem statement; may include inline \LaTeX{} for equations.\\
        \texttt{hint} & String & Optional hint for the benchmark instruction prompt.\\
        \texttt{step} & Array[String] & Reasoning steps to solve the problem.\\
        \texttt{final} & String & Ground-truth target answer for grading. \\
        \texttt{type} & String & Expected output type for grading: "\texttt{numeric}", "\texttt{symbolic}", or "\texttt{textual}".\\
        \texttt{meta} & Dictionary & Metadata (e.g.,  \texttt{year}, \texttt{author}, \texttt{source}).\\
        \bottomrule
    \end{tabular}
\end{table}

\subsection{Authors\label{appendix:lwsauthors}}

\begin{table}[H]
\centering
\caption{Authors of LWS Summer School Solution Sets and the number of questions within RWS.}
\begin{tabular}{@{}lc@{}}
\toprule
\textbf{Author Name}           & \textbf{Number of Questions} \\ \midrule
Vytenis Vasyliunas    & 29                  \\
Martin Lee            & 26                  \\
Karel Schrijver       & 17                  \\
Fran Bagenal          & 17                  \\
Merav Opher           & 10                  \\
Matthias Rempel       & 10                  \\
Mark Miesch           & 9                   \\
J. R. Jokipii         & 8                   \\
Timothy Fuller-Rowell & 8                   \\
Sabine Stanley        & 6                   \\
Justin Kasper         & 6                   \\
Kevin Forbes          & 5                   \\
J. T. Gosling         & 4                   \\
Rachel Osten          & 3                   \\ \bottomrule
\end{tabular}
\end{table}

\clearpage

\section{Multi-Agent System Architectures}
\label{appendix:mas}

This appendix summarizes the four multi-agent design patterns we evaluated. For each pattern, we outline the control flow, the data and interface contracts, and the revision policy. We also provide three diagrams for each pattern: an agentic workflow diagram, a UML activity diagram, and a UML sequence diagram. We use the Google Agent Development Kit (ADK) \cite{noauthor_agent_nodate} to set up these agent design patterns for the multi-agent system benchmark. The complete instruction prompts for each agent role are documented in \appendixautorefname{}~\ref{appendix:instructions}.

\begin{figure}[H]
  \centering
  \includegraphics[width=0.6\linewidth]{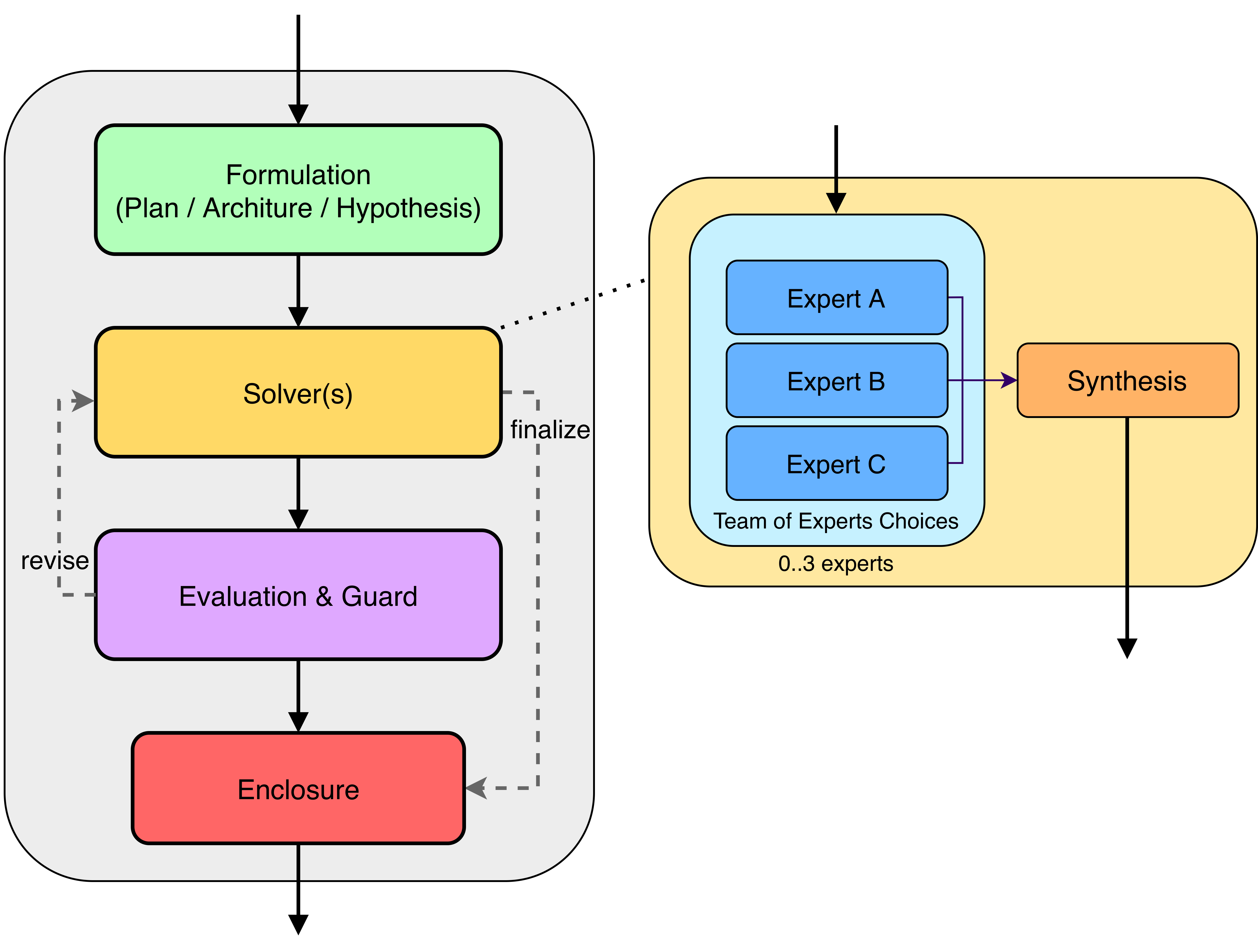}
  \caption{General Design Philosophy of Multi-Agent System.}
  \label{fig:mas-general}
\end{figure}

Our general design philosophy of these multi-agent systems is shown in Figure~\ref{fig:mas-general}.

\subsection{Systems-engineering-of-Thoughts Agentic Reasoning (STAR)}
\label{appendix:star}

Systems engineering evolved from solving complex integration challenges in mission-critical domains such as spacecraft payloads, guidance and navigation systems, and large-scale infrastructure projects, where failures carry substantial cost and complex interactions must be coordinated through well-characterized components and interfaces. Foundational references such as the NASA Systems Engineering Handbook and the INCOSE Systems Engineering Handbook \cite{hirshorn_nasa_2017,incose_handbook_2023} emphasize requirements, interface management, and verification and validation throughout the project life cycle. Case studies from aerospace and automotive programs (e.g., Lockheed Martin, Northrop Grumman, and General Motors) link high mission success and product reliability to disciplined systems engineering and Model-Based Systems Engineering (MBSE) practices \cite{dean_model-based_2015,dove_case_2017,dambrosio_mbse_2019}.

By contrast, LLM-based workflows are often treated as monolithic black boxes: users initiate natural-language prompts and expect the model’s “thought process” to yield correct answers. This is unreliable, both because LLMs can produce convincing but factually incorrect content (``hallucinations'') and because their behavior is sensitive to how the prompt is phrased \cite{farquhar_detecting_2024,kalai_why_2025}. Recent work shows that vague or underspecified instructions degrade LLM performance \cite{yang_what_2025}, whereas well-specified step-by-step instructions and explicit problem decompositions can substantially improve reasoning accuracy \cite{wei_chain--thought_2023}. When inputs are ambiguous, prompting LLMs to ask clarification questions leads to better answers than forcing an immediate response \cite{kuhn_clam_2023}. Robustness analyses further indicate that natural input noise, such as spelling errors or small text corruptions, can significantly reduce accuracy \cite{aliakbarzadeh_exploring_2025}. These findings motivate more systematic design and checking of agent instructions, architectures, and intermediate reasoning steps, rather than ad hoc, one-shot prompts.

We adopt this perspective through our \emph{Systems-engineering-of-Thoughts Agentic Reasoning (STAR)} method. Under STAR, the LLM’s ``thought process'' is not an opaque transcript from a single call, but an engineered process with well-defined modules, contracts, and checkpoints. Each agent role has a clearly defined responsibility, communicates through constrained interfaces (e.g., schemas, unit conventions, and output formats in our benchmark), and operates under explicit evaluation and revision rules, echoing MBSE’s emphasis on model and interface traceability \cite{hart_mbse_2015}.

Concretely, we instantiate a STAR-inspired architecture with four macro-modules and an optional team-of-experts branch within the solver stage as shown in Figure~\ref{fig:mas-general}:

\textbf{Formulation} proposes a high-level plan or hypothesis based on user queries, specifications, and requirements. In our implementation, this includes formalizing assumptions, units, and checks before queries are sent to the solver pipeline.

\textbf{Solver(s)} act on the plan, performing solution steps to provide (candidate) solutions. For RWS instances, this includes deriving algebraic expressions, propagating units, and computing numerical values, as specified in the solver instructions.

\textbf{Evaluation \& Guard} check internal consistency (units, formatting, task-specific constraints) and decide whether candidate solutions should be revised or approved and passed downstream.

\textbf{Enclosure} converts approved solutions into the required format, matching the benchmark schema so that the output is usable for programmatic grading and downstream applications.

\textbf{Team of Experts Choices \& Synthesis} (optional, within the Solver stage) instantiates a limited number of specialized expert agents (e.g., physics, mathematics, or coding experts) based on the plan from Formulation. Their outputs are then combined by a synthesis step before returning a unified solution to the main pipeline.

This architecture is intended as an example of how a systems-engineering perspective can be realized in practice. Other STAR-based designs may choose different ways of modularizing agent roles and communication patterns.

STAR is designed to complement existing reasoning approaches such as chain-of-thought, tree-of-thoughts, and graph-of-thoughts \cite{wei_chain--thought_2023,yao_tree_2023,besta_graph_2024}. Although our current experiments use relatively compact prompt structures within each module for initial analysis, STAR provides an architectural layer that can host these approaches by specifying how such subsystems are channeled and coordinated.

Beyond this specific implementation, STAR serves as a design template for future LLM-based agentic architectures. Systems developed under STAR should emphasize rigor in defining role modules, interface responsibilities, instruction formats, and guard conditions. When appropriate, prompting schemes can adopt the philosophy of MBSE at the system level \cite{estefan_mbse_2008}, with explicit representations of requirements, interfaces, states, and verification checks, while still preserving the prompting flexibility \cite{hart_mbse_2015} by general users. This systems-engineering perspective turns loosely specified agent instructions into clear and analyzable pipelines, facilitating the routing of complex logical and algebraic problems to appropriate modules and the use of LLMs as reasoning engines within larger scientific and mission-critical environments.

\subsection{HMAW (Hierarchical Multi-Agent Workflow: CEO $\rightarrow$ Manager $\rightarrow$ Worker)}
\label{appendix:hmaw-arch}

HMAW follows a fixed, top-down, single-pass pipeline based on prior Hierarchical Multi-Agent Workflow research \cite{liu_towards_2025}. We use it as a simple hierarchical baseline to compare with other multi-agent patterns through its straightforward communication architecture for agent collaboration study.

In this system, the Conductor agent holds the user query and invokes a sequence of role agents: CEO, Manager, and Worker. The CEO creates a first message packet (MP1) that contains actionable instructions for the Manager. The Manager translates MP1 into a second message packet (MP2) with concrete execution steps for the Worker. Finally, the Worker executes MP2 and returns the final answer. This hierarchical structure does not involve retries or branching.

\subsubsection*{HMAW Diagrams}

\begin{figure}[H]
  \centering
  \includegraphics[width=0.6\linewidth]{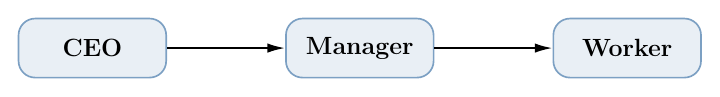}
  \caption{HMAW: Agentic Workflow Diagram.}
  \label{fig:hmaw}
\end{figure}

\begin{figure}[H]
  \centering
  \includegraphics[width=0.9\linewidth]{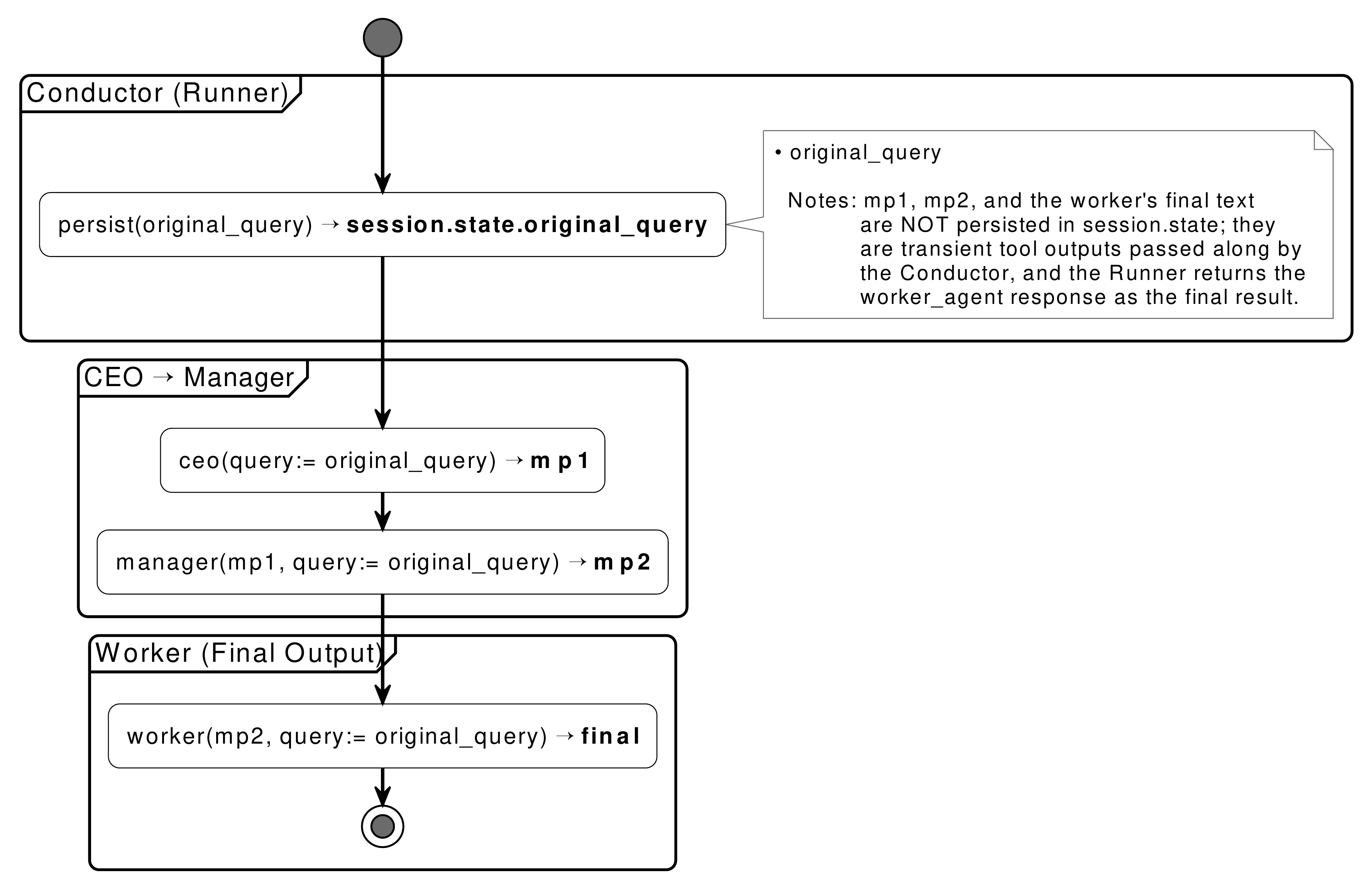}
  \caption{HMAW: UML Activity Diagram (Google ADK).}
  \label{fig:hmaw-act}
\end{figure}

\begin{figure}[H]
  \centering
  \includegraphics[width=\linewidth]{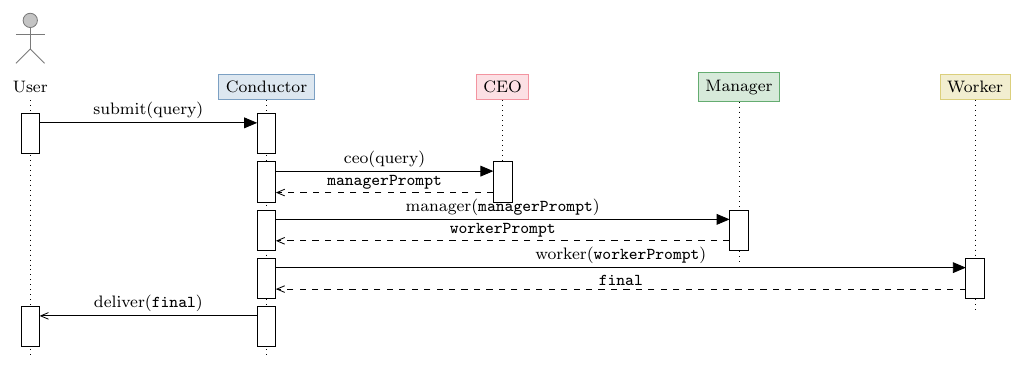}
  \caption{HMAW: UML Sequence Diagram.}
  \label{fig:hmaw-seq}
\end{figure}

\subsection{PACE (Plan $\rightarrow$ Answer $\rightarrow$ Critique $\rightarrow$ Enclose)}
\label{appendix:pace-arch}

PACE is a streamlined pipeline with a single bounded retry. It aims to balance reliability, simplicity, and efficiency. The self-verification and self-critique mechanisms \cite{gou_critic_2024,lee_revise_2025,tan_improved_2025,shinn_reflexion_2023} of the system often improve the reliability of structured math-style problems by detecting common errors without introducing heavy orchestration.

In this system, the Conductor agent stores the user query and calls four role agents in order: Plan, Answer, Critique, and Enclose. The Planner agent creates a minimal plan that outlines the requirements needed to address the query, including any checks. The Answer agent follows this plan and produces a reasoning proposal along with the answer. The Critique agent assesses the proposed answer based on acceptance, confidence, and issues. If the Critique agent rejects the answer, it offers a fix, and the Answer agent then gets a retry. Finally, the Encloser agent packages the final answer into the necessary output format.

\subsubsection*{PACE Diagrams}

\begin{figure}[H]
  \centering
  \includegraphics[width=0.8\linewidth]{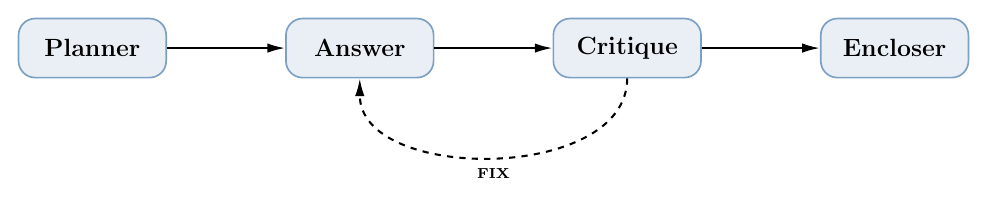}
  \caption{PACE: Agentic Workflow Diagram.}
  \label{fig:pace}
\end{figure}

\begin{figure}[H]
  \centering
  \includegraphics[width=\linewidth]{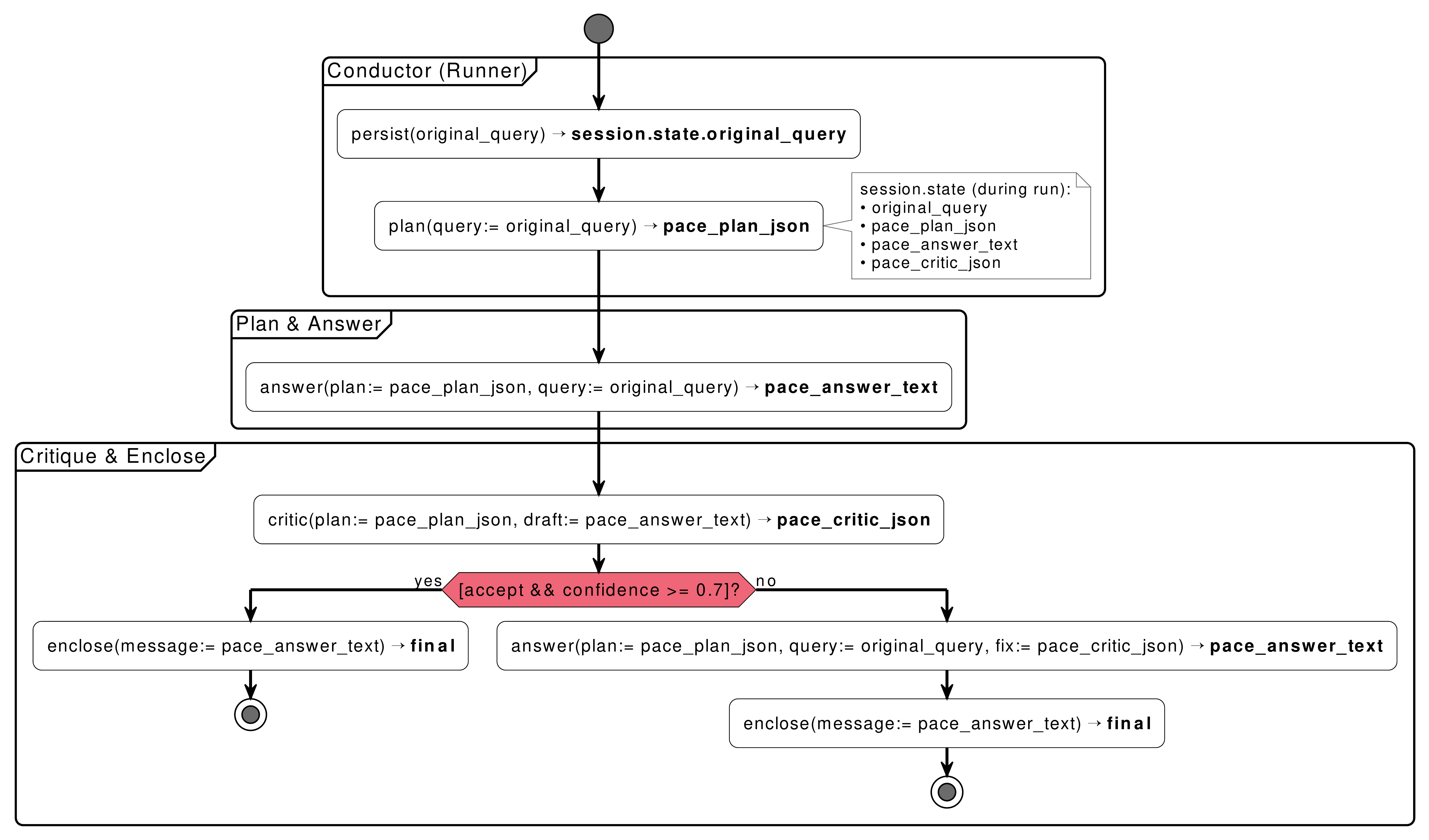}
  \caption{PACE: UML Activity Diagram (Google ADK).}
  \label{fig:pace-act}
\end{figure}

\begin{figure}[H]
  \centering
  \includegraphics[width=\linewidth]{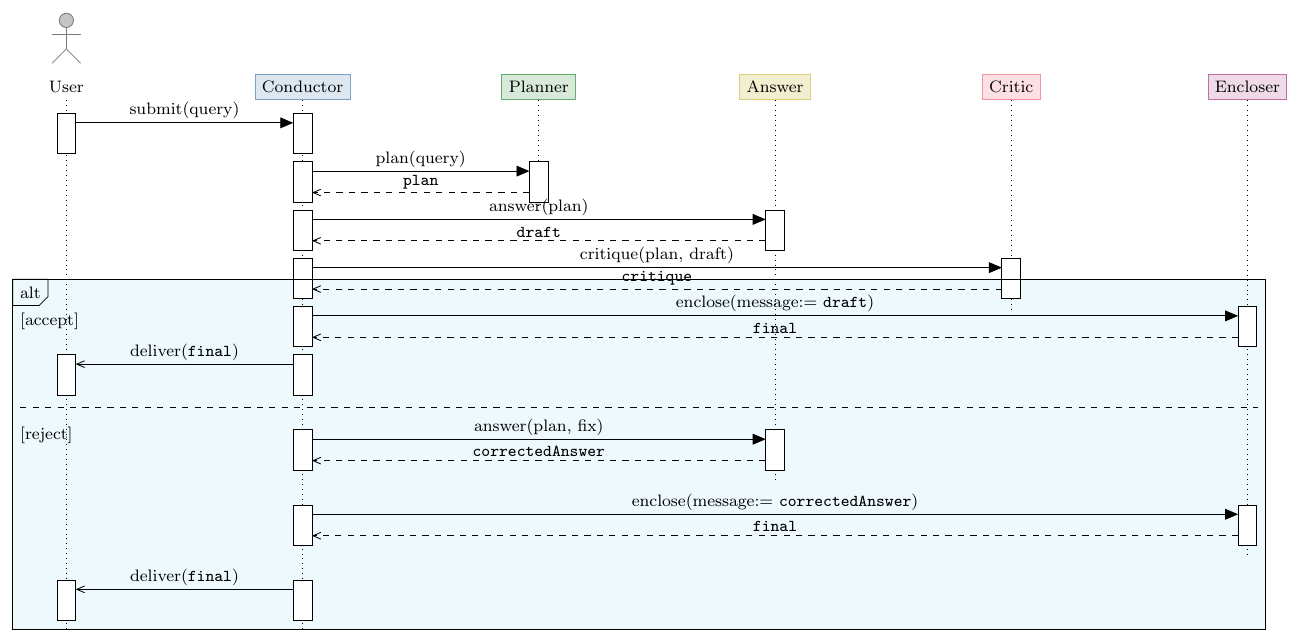}
  \caption{PACE: UML Sequence Diagram.}
  \label{fig:pace-seq}
\end{figure}

\subsection{PHASE (Plan $\rightarrow$ Hypothesize $\rightarrow$ Analyze $\rightarrow$ Solve $\rightarrow$ Evaluate $\rightarrow$ Finalize)}
\label{appendix:phase-arch}

PHASE is a physics- and math-aware pipeline that formulates assumptions and required units before attempting a full solution. It takes inspiration from SiriuS \cite{zhao_sirius_2025} and prior work that views scientific reasoning as a step-by-step process. This process starts with the formation of a hypothesis \cite{shen_satori_2025}, moves to targeted verification, and finalizes with correction \cite{lee_revise_2025,gou_critic_2024}.

In this system, the Conductor agent tracks the user query and goes through six stages: Plan, Hypothesize, Analyze, Solve, Evaluate, and Finalize. The Planner agent creates a clear requirements plan that includes at most two explicit checks. The Hypothesizer agent produces a compact hypothesis pack that lists assumptions of what is known and unknown, any relevant equations, and required units. The Analyzer agent performs a brief derivation or reasoning step based on that hypothesis and suggests a possible approach. The Solver agent then returns the final answer. The Evaluator agent examines the final answer; it states whether the answer is accepted, gives a confidence score, and identifies issues along with a suggested fix. If the Evaluator rejects the answer, one bounded correction is sent back to the Solver. The Finalizer agent then puts the accepted answer in the required output format.

\subsubsection*{PHASE Diagrams}

\begin{figure}[H]
  \centering
  \includegraphics[width=\linewidth]{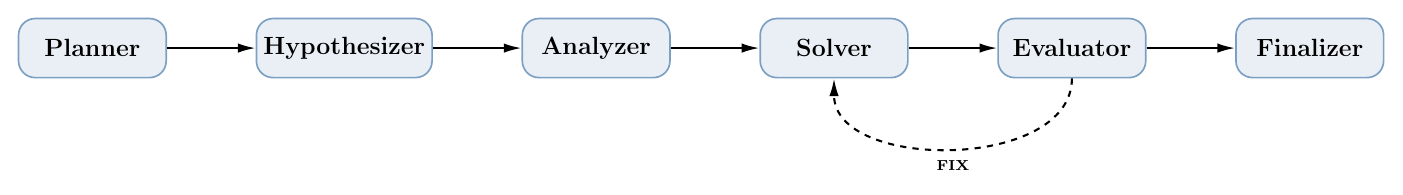}
  \caption{PHASE: Agentic Workflow Diagram.}
  \label{fig:phase}
\end{figure}

\begin{figure}[H]
  \centering
  \includegraphics[width=\linewidth]{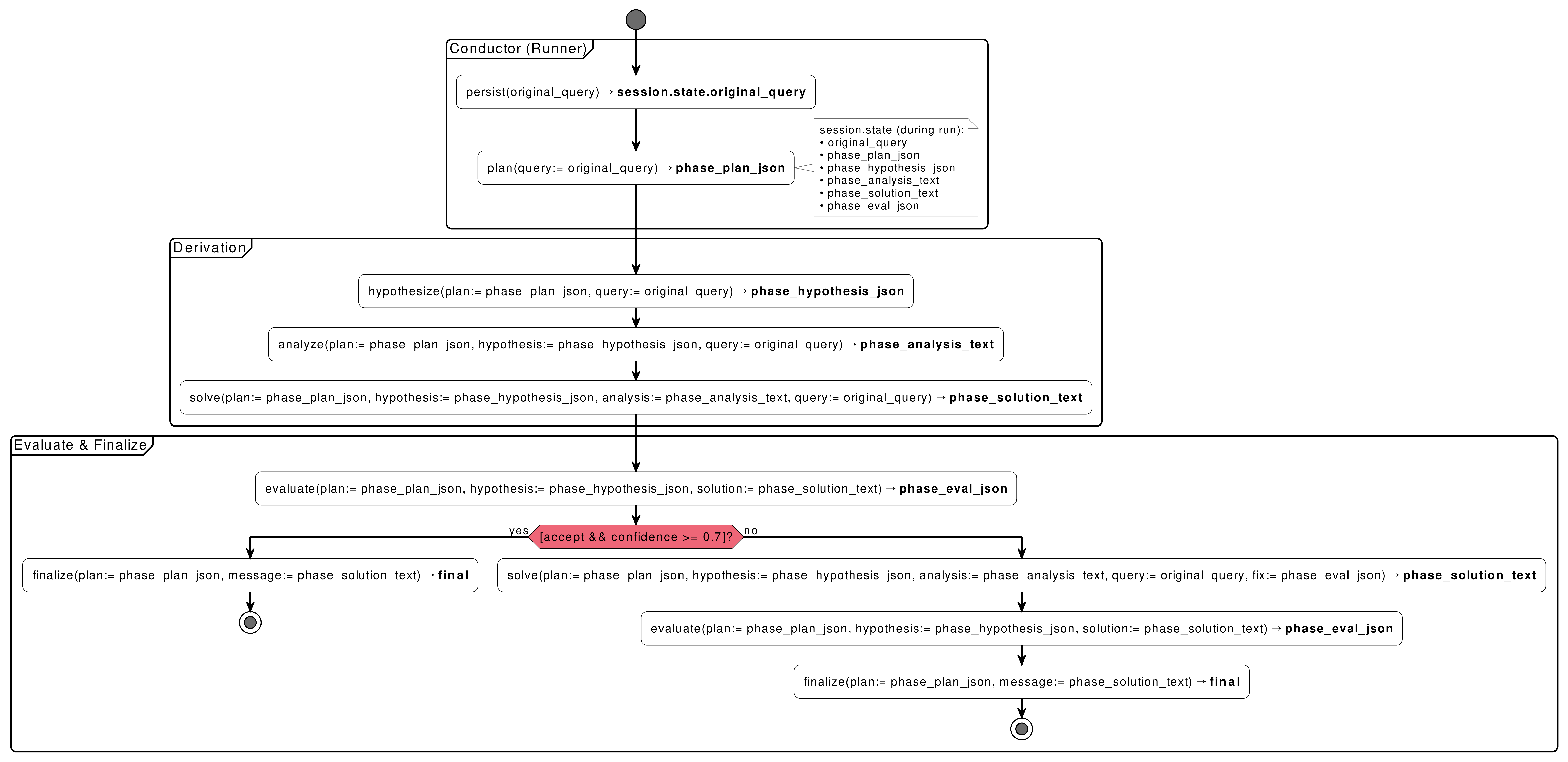}
  \caption{PHASE: UML Activity Diagram (Google ADK).}
  \label{fig:phase-act}
\end{figure}

\begin{figure}[H]
  \centering
  \includegraphics[width=\linewidth]{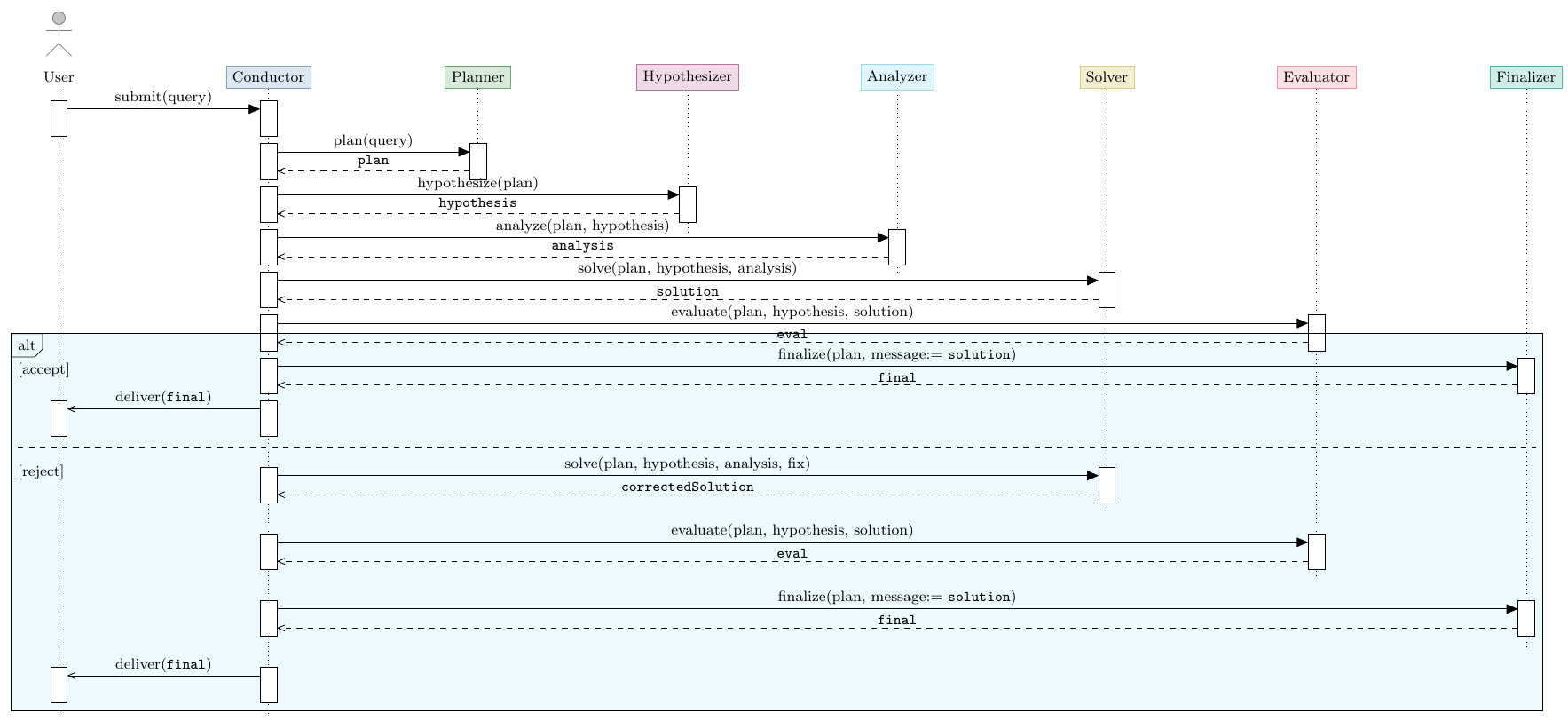}
  \caption{PHASE: UML Sequence Diagram.}
  \label{fig:phase-seq}
\end{figure}

\subsection{SCHEMA (Systems-Engineering Coordinated Hierarchical Expert Multi-Agent)}
\label{appendix:schema-arch}

SCHEMA is a design pattern inspired by systems engineering principles \cite{hirshorn_nasa_2017} that dynamically forms a team of experts for each task. It uses Model-Based Systems Engineering (MBSE) practices from space applications and ideas from self-evolving multi-agent systems like MAS-ZERO \cite{ke_mas-zero_2025} and related role-allocation work \cite{zhao_sirius_2025}. The aim is to ensure clear requirement tracking, interface control, and verification checkpoints instead of relying on a single monolithic prompt.

In this system, the Conductor agent stores the user query and calls the following roles: Architect, Allocator, one or more Experts, a Synthesizer, a Guard, and a Finalizer. The Architect agent creates an architecture JSON that details the type of question, format hints, acceptance checks, and an expert sequence with role assignments; it also defines interface contracts. The Allocator agent standardizes those roles, enforces interface constraints, and provides specific plans to each Expert. Each Expert then works in sequence, providing a reasoning trace and a candidate output. The Synthesizer agent collects and coordinates these expert outputs and sends a unified answer to the Guard agent. The Guard agent evaluates the synthesized answer against the stated requirements; if it finds an issue, it sends a minimal correction back to the Synthesizer. Finally, the Finalizer packages the accepted result in the required output format.

\subsubsection*{SCHEMA Diagrams}

\begin{figure}[H]
  \centering
  \includegraphics[width=\linewidth]{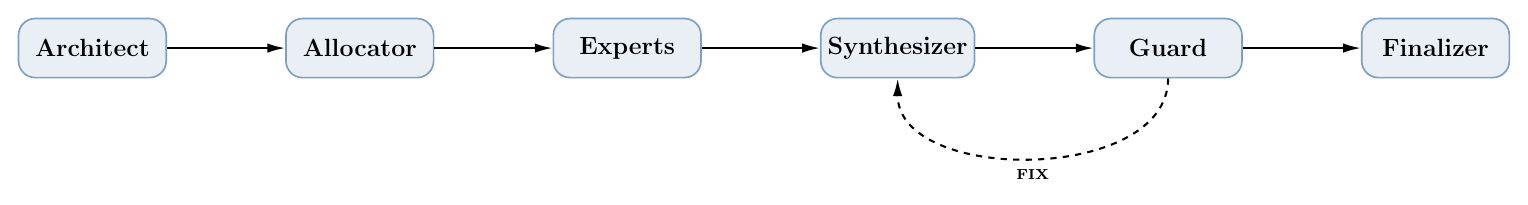}
  \caption{SCHEMA: Agentic Workflow Diagram.}
  \label{fig:schema}
\end{figure}

\begin{figure}[H]
  \centering
  \includegraphics[width=0.88\linewidth]{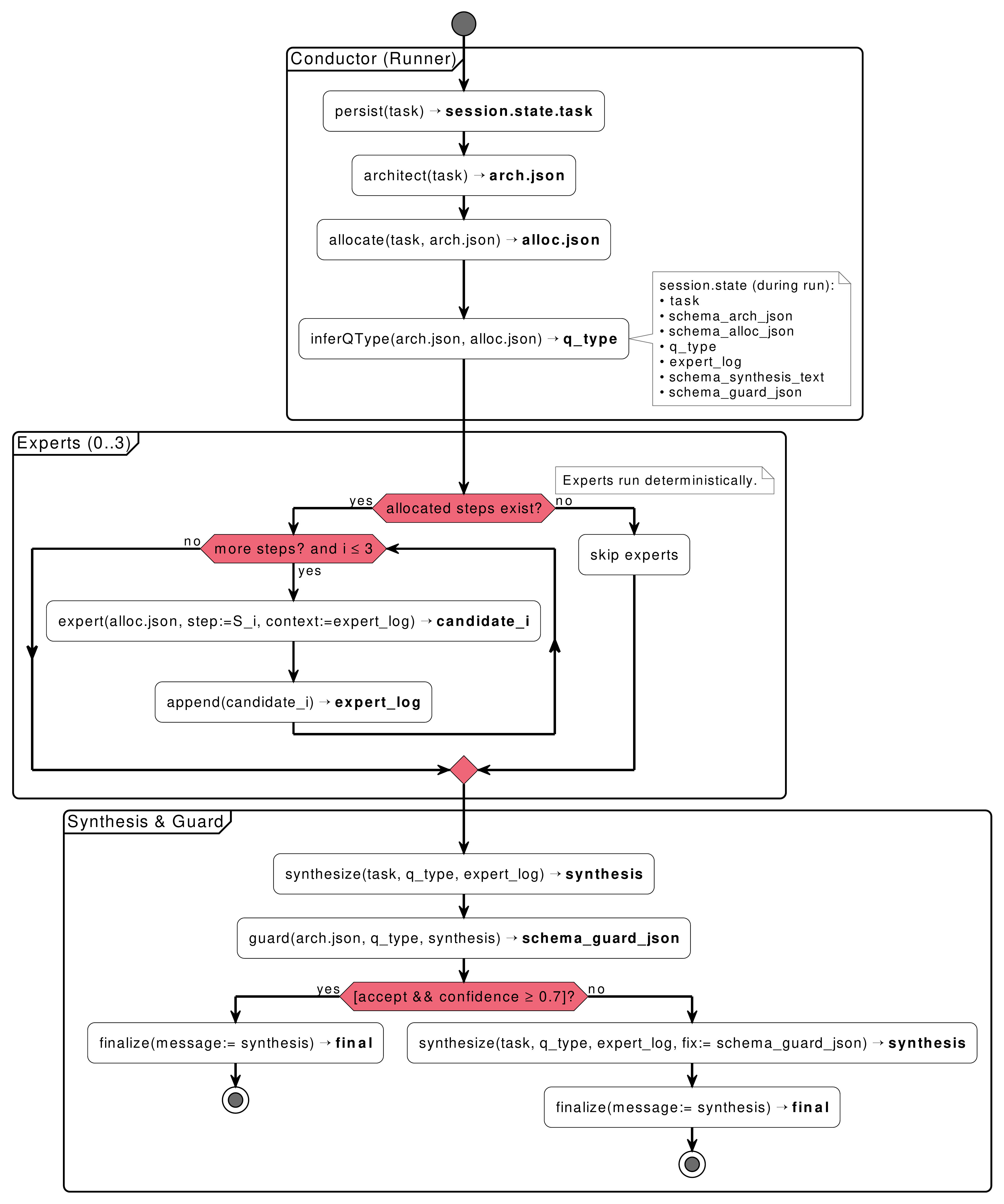}
  \caption{SCHEMA: UML Activity Diagram (Google ADK).}
  \label{fig:schema-act}
\end{figure}

\begin{figure}[H]
  \centering
  \includegraphics[width=\linewidth]{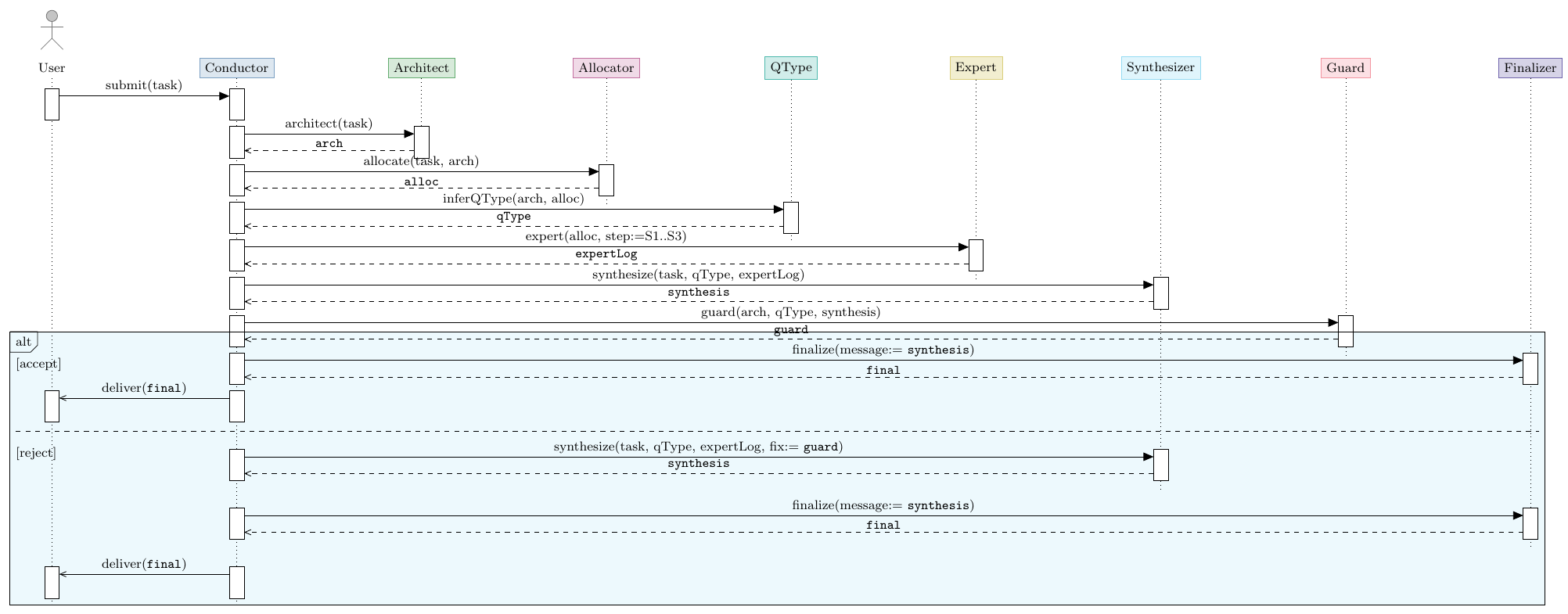}
  \caption{SCHEMA: UML Sequence Diagram.}
  \label{fig:schema-seq}
\end{figure}

\clearpage

\section{Multi-Agent Systems Instruction Prompts}
\label{appendix:instructions}
This appendix lists the instruction prompts used by each multi-agent configuration in the Google Agent Development Kit (ADK) \cite{noauthor_agent_nodate}. For every system, we provide both the system-level instructions and the role-specific instructions for each agent.

Each agent is defined by its responsibility within the workflow, the interface contract that outlines what it must receive and what it must output, and any constraints it must enforce.

\subsection{HMAW Instruction Prompts \cite{liu_towards_2025}}

\begin{agentinstruction}[title=HMAW: System (Conductor Agent)]
You are a master orchestrator for a hierarchical multi-agent workflow (HMAW).\\
Your goal is to take a user's query and initiate a CEO -> Manager -> Worker pipeline.\\
\\
You MUST follow this sequence of steps precisely:\\
\\
1. **Store Query**: When you receive the user's query in `\{new\_message.text\}`, you MUST \\
\promptspace{}\promptspace{}\promptspace{}first store it in the session state by setting `session.state.original\_query`.\\
\\
2. **Invoke CEO**: Call the `ceo\_agent` tool with the original user query (`\{session.state.original\_query\}`) \\
\promptspace{}\promptspace{}\promptspace{}as the `query` parameter.\\
\\
3. **Invoke Manager**: Take the output from the `ceo\_agent` tool and call the `manager\_agent` \\
\promptspace{}\promptspace{}\promptspace{}tool with this output as the `query` parameter. The original query is available in the session state.\\
\\
4. **Invoke Worker**: Take the output from the `manager\_agent` tool and call the `worker\_agent` tool with \\
\promptspace{}\promptspace{}\promptspace{}this output as the `query` parameter. The original query is available in the session state.\\
\\
Your job is complete after you have invoked the worker\_agent. Do not wait for a response from the worker. Do not produce a final response.
\end{agentinstruction}

\begin{agentinstruction}[title=HMAW: CEO Agent]
**Your ROLE**: <CEO>\\
**Description**: You are the CEO of an entirely LLM-based company where all employees are LLMs. The \\
\promptspace{}\promptspace{}\promptspace{}company's goal is to generate the best possible response tailored to the user's request.\\
**Company Structure**: CEO (LLM) -> MANAGER (LLM) -> WORKER (LLM) -> USER\\
**Company Workflow**:\\
1. The CEO receives the input (prompt P) from the human user.\\
2. The CEO generates detailed instructions (prompt MP1) for the MANAGER LLM.\\
3. According to MP1, the MANAGER then creates detailed instructions (prompt MP2) for the WORKER \\
\promptspace{}\promptspace{}\promptspace{}\promptspace{}LLM.\\
4. The WORKER LLM uses MP2 to generate the golden response (Output O) for the user.\\
**IMPORTANT**:\\
- As the CEO, your task is to generate the prompt MP1 for the MANAGER LLM so that the MANAGER \\
\promptspace{}\promptspace{}LLM can generate the golden prompt (MP2) for the WORKER LLM. The final goal is to make the output\\
\promptspace{}\promptspace{}(O) of the WORKER LLM highly tailored, pleasing, and accurate.\\
- As the CEO, do not output anything else; only provide the prompt MP1 to the MANAGER LLM.\\
- As the CEO, do not try to generate the final output for the user. This will be done by the WORKER LLM \\
\promptspace{}\promptspace{}supervised by the MANAGER LLM.\\
- If you need to repeat the human user's input, repeat it exactly without any placeholders.\\
- Begin your response with **Detailed Instructions to MANAGER**:
\end{agentinstruction}

\begin{agentinstruction}[title=HMAW: Manager Agent]
**Your ROLE**: <MANAGER>\\
**Description**: You are the MANAGER in an entirely LLM-based company where all employees are \\
\promptspace{}\promptspace{}\promptspace{}\promptspace{}LLMs. The company's goal is to generate the best possible response tailored to the user's request.\\
**Company Structure**: CEO (LLM) -> MANAGER (LLM) -> WORKER (LLM) -> USER\\
**Company Workflow**:\\
1. The CEO receives the input (prompt P) from the human user.\\
2. The CEO generates detailed instructions (prompt MP1) for the MANAGER LLM.\\
3. According to MP1, the MANAGER then creates detailed instructions (prompt MP2) for the WORKER \\
\promptspace{}\promptspace{}\promptspace{}\promptspace{}LLM.\\
4. The WORKER LLM uses MP2 to generate the golden response (Output O) for the user.\\
**IMPORTANT**:\\
- As the MANAGER, your task is to generate the prompt MP2 for the WORKER LLM so that the WORKER \\
\promptspace{}\promptspace{}can provide the golden response to the user according to MP2.\\
- Do not output anything else; only provide the prompt MP2 to the WORKER LLM.\\
- Do not try to generate the final output for the user; this will be done by the WORKER using the prompt \\
\promptspace{}\promptspace{}generated by you.\\
- If you need to repeat the human user's input, repeat it exactly without placeholders.\\
- Begin your response with **Detailed Instructions to WORKER**:
\end{agentinstruction}

\begin{agentinstruction}[title=HMAW: Worker Agent]
**Your ROLE**: <WORKER>\\
**Description**: You are the WORKER in an entirely LLM-based company where all employees are LLMs. The company's goal is to generate the best possible response tailored to the user's request.\\
**Company Structure**: CEO (LLM) -> MANAGER (LLM) -> WORKER (LLM) -> USER\\
**Company Workflow**:\\
1. The CEO receives the input (prompt P) from the human user.\\
2. The CEO generates detailed instructions (prompt MP1) for the MANAGER LLM.\\
3. According to MP1, the MANAGER then creates detailed instructions (prompt MP2) for the WORKER \\
\promptspace{}\promptspace{}\promptspace{}\promptspace{}LLM.\\
4. The WORKER LLM uses MP2 to generate the golden response (Output O) for the user.\\
**IMPORTANT**:\\
- As the WORKER, your task is to generate the final output (O) for the user with the prompt from the \\
\promptspace{}\promptspace{}MANAGER. This output should be highly tailored, pleasing, and accurate.\\
- Ensure the response is excellent and directly talking to the user.\\
- Do not say you cannot answer.
\end{agentinstruction}

\subsection{PACE Instruction Prompts}

\begin{agentinstruction}[title=PACE: System (Conductor Agent)]
You are the **PACE Conductor**. You orchestrate tools and never address the user.\\
\\
Do exactly this sequence:\\
\\
1. Persist the user's query:\\
- Set session.state.original\_query = \{new\_message.text\}\\
\\
2. Call 'pace\_planner' with:\\
query = "QUERY:\textbackslash{}n\{session.state.original\_query\}"\\
\\
3. Call 'pace\_answer' with:\\
query = "QUERY:\textbackslash{}n\{session.state.original\_query\}\textbackslash{}n\textbackslash{}nPLAN:\textbackslash{}n\{session.state.pace\_plan\_json\}"\\
\\
4. Call 'pace\_critic' with:\\
query = "PLAN:\textbackslash{}n\{session.state.pace\_plan\_json\}\textbackslash{}n\textbackslash{}nDRAFT:\textbackslash{}n\{session.state.pace\_answer\_text\}"\\
\\
5. If the critic JSON indicates accept==true AND confidence >= 0.7:\\
- Call 'pace\_encloser' with:\\
query = "MESSAGE:\textbackslash{}n\{session.state.pace\_answer\_text\}"\\
- STOP after the tool response.\\
\\
Otherwise (reject case):\\
- Call 'pace\_answer' again with:\\
query = "QUERY:\textbackslash{}n\{session.state.original\_query\}\textbackslash{}n\textbackslash{}nPLAN:\textbackslash{}n\\
\{session.state.pace\_plan\_json\}\textbackslash{}n\textbackslash{}nFIX:\textbackslash{}n\{session.state.pace\_critic\_json\}"\\
- Then call 'pace\_encloser' with:\\
query = "MESSAGE:\textbackslash{}n\{session.state.pace\_answer\_text\}"\\
- STOP after the tool response.\\
\\
IMPORTANT:\\
- Do not output a final user message yourself. Your job ends after invoking the encloser.\\
- Keep any intermediate strings as-is (no extra formatting or fences).\\
\\
\end{agentinstruction}
\begin{agentinstruction}[title=PACE: Planner Agent]
You are the **PACE Planner**.\\
\\
GOAL:\\
- Analyze the original user query and draft a minimal executable plan the downstream agents can follow.\\
- Do not perform calculations; produce only the plan.\\
\\
INPUT:\\
- Query: \{session.state.original\_query\}\\
\\
OUTPUT:\\
- Return a EXACTLY one JSON report (no prose, no fences), with keys:\\
\{\\
\promptspace{}\promptspace{}"expected\_type": "<numeric|symbolic|textual>",\\
\promptspace{}\promptspace{}"required\_units": "<unit string if the query specifies it (e.g., 'nT', 'cm\^{}-3'); else null>",\\
\promptspace{}\promptspace{}"format\_hints": "<optional formatting notes (e.g., required symbolic form); else empty string>",\\
\promptspace{}\promptspace{}"plan": ["<step 1>", "<step 2>", "<step 3?>", "<step 4?>"],\\
\promptspace{}\promptspace{}"checks": ["<acceptance check 1>", "<acceptance check 2?>"]\\
\}\\
\\
RULES:\\
- If the query says “ONLY ...” (e.g., ONLY numeric, ONLY symbolic, ONLY textual), set expected\_type \\
\promptspace{}\promptspace{}accordingly.\\
- If the query specifies an output unit (e.g., nT, km/s, cm\^{}-3), copy it into required\_units verbatim; otherwise \\
\promptspace{}\promptspace{}set required\_units to null.\\
- If the query implies a specific symbolic form or textual constraint, summarize it in format\_hints; \\
\promptspace{}\promptspace{}otherwise use "".\\
- Keep steps actionable and atomic; avoid re-stating the user’s question; no calculations.\\
- Keep plan length \texttt{<=} 4 steps and checks length \texttt{<=} 2.\\
\end{agentinstruction}

\begin{agentinstruction}[title=PACE: Answer Agent]
You are the **PACE Answer Agent**.\\
\\
You receive a packed message that includes:\\
- QUERY: <original user query text>\\
- PLAN: <JSON from the Planner>  // includes expected\_type, required\_units, optional format\_hints\\
- (Optional) FIX: <brief targeted fix instruction from the Critic>\\
\\
RULES:\\
- Follow the plan to produce a concise solution. Keep reasoning tight. Obey “ONLY …” constraints.\\
- If numeric data are insufficient or unspecified, prefer a symbolic/LaTeX result rather than guessing.\\
- If PLAN.required\_units is non-null, convert to those units and include them on the Final Answer line.\\
- If PLAN.format\_hints specifies a required symbolic form or textual constraint, follow it.\\
\\
OUTPUT (plain text):\\
Reasoning:\\
<up to 4 short lines (bullets or terse sentences). Do not include full derivations; keep to essential steps or substitutions.>\\
\\
FINAL ANSWER:\\
<one line final answer only; match 'expected\_type' exactly; keep units/LaTeX canonical>\\
- Do NOT include labels like 'symbolic:', 'textual:', or 'numeric:'.\\
- symbolic: a single inline math expression (e.g., $...$); no surrounding prose.\\
- textual: one concise sentence.\\
- numeric: number + (required) units; use SI by default if units not specified.\\
\\
\end{agentinstruction}

\begin{agentinstruction}[title=PACE: Critic Agent]
You are the **PACE Critic**.\\
\\
INPUT is a packed message:\\
- PLAN: <planner JSON>\\
- DRAFT: <answer output>\\
\\
Evaluate ONLY the Final Answer line in DRAFT against:\\
- PLAN.expected\_type (type conformity)\\
- PLAN.required\_units (if provided; otherwise SI is acceptable)\\
- PLAN.checks (\texttt{<=} 2 acceptance checks)\\
- Basic sanity: dimensions/units consistency, LaTeX syntactic validity.\\
\\
OUTPUT:\\
- Return a strict JSON report (no prose, no fences), with keys:\\
\promptspace{}\promptspace{}\{\\
\promptspace{}\promptspace{}\promptspace{}\promptspace{}"accept": <true\_or\_false>,\\
\promptspace{}\promptspace{}\promptspace{}\promptspace{}"confidence": <float 0..1>,\\
\promptspace{}\promptspace{}\promptspace{}\promptspace{}"issues": ["<issue 1>", "<issue 2?>"],\\
\promptspace{}\promptspace{}\promptspace{}\promptspace{}"fix": "<brief targeted fix instruction>"\\
\promptspace{}\promptspace{}\}\\
\\
RULES:\\
- confidence: use \textasciitilde{}0.9 if all checks pass cleanly; \textasciitilde{}0.6 for minor format fixes; \texttt{<=} 0.4 for substantive errors.\\
- numeric: ensure a real number (with optional exponent) and proper unit token; allow +/-5\% tolerance unless the query states otherwise.\\
- symbolic: exactly one inline expression (between $...$); no words; simplified if possible.\\
- textual: keep to one concise sentence.\\
\\
\end{agentinstruction}

\begin{agentinstruction}[title=PACE: Encloser Agent]
You are the **PACE Encloser**.\\
\\
INPUT is a packed message:\\
- MESSAGE: <the latest Answer/Refiner output>\\
\\
ALWAYS produce this exact format:\\
Reasoning:\\
<extract and compress to <\texttt{<=} 3 short lines; if none provided, synthesize a minimal 1-2 line summary>\\
\\
ENSURE the last lines are:\\
Final Answer:\\
\promptspace{}\promptspace{}<canonical final answer only on this line>\\
\\
RULES:\\
- Do NOT wrap content in backticks or code fences.\\
- For symbolic, output a single inline expression; no extra text.\\
- Remove any residual labels (“symbolic:”, “textual:”, “numeric:”).\\
- Never add additional commentary after the Final Answer line.\\
\\
\end{agentinstruction}

\subsection{PHASE Instruction Prompts}

\begin{agentinstruction}[title=PHASE: System (Conductor Agent)]
You are the **PHASE Conductor**. You orchestrate tools and never address the user.\\
\\
Do exactly this sequence:\\
\\
1. Set session.state.original\_query = \{new\_message.text\}\\
\\
2. Call 'phase\_planner' with:\\
query = "QUERY:\textbackslash{}n\{session.state.original\_query\}"\\
\\
3. Call 'phase\_hypothesizer' with:\\
query = "QUERY:\textbackslash{}n\{session.state.original\_query\}\textbackslash{}n\textbackslash{}nPLAN:\textbackslash{}n\{session.state.phase\_plan\_json\}"\\
\\
4. Call 'phase\_analyzer' with:\\
query = "QUERY:\textbackslash{}n\{session.state.original\_query\}\textbackslash{}n\textbackslash{}nPLAN:\textbackslash{}n\{session.state.phase\_plan\_json\}\\
\textbackslash{}n\textbackslash{}nHYPOTHESIS:\textbackslash{}n\{session.state.phase\_hypothesis\_json\}"\\
\\
5. Call 'phase\_solver' with:\\
query = "QUERY:\textbackslash{}n\{session.state.original\_query\}\textbackslash{}n\textbackslash{}nPLAN:\textbackslash{}n\{session.state.phase\_plan\_json\}\\
\textbackslash{}n\textbackslash{}nHYPOTHESIS:\textbackslash{}n\{session.state.phase\_hypothesis\_json\}\textbackslash{}n\textbackslash{}nANALYSIS:\\
\textbackslash{}n\{session.state.phase\_analysis\_text\}"\\
\\
6. Call 'phase\_evaluator' with:\\
query = "PLAN:\textbackslash{}n\{session.state.phase\_plan\_json\}\textbackslash{}n\textbackslash{}nHYPOTHESIS:\textbackslash{}n\\
\{session.state.phase\_hypothesis\_json\}\textbackslash{}n\textbackslash{}nSOLUTION:\textbackslash{}n\{session.state.phase\_solution\_text\}"\\
\\
7. Gate (single bounded retry):\\
- Parse \{session.state.phase\_eval\_json\} for accept and confidence.\\
- If accept==true AND confidence >= 0.7:\\
* Call 'phase\_finalizer' with:\\
query = "PLAN:\textbackslash{}n\{session.state.phase\_plan\_json\}\textbackslash{}n\textbackslash{}nMESSAGE:\textbackslash{}n\\
\{session.state.phase\_solution\_text\}"\\
* STOP after the tool response.\\
- Otherwise (reject case):\\
* Call 'phase\_solver' again with:\\
query = "QUERY:\textbackslash{}n\{session.state.original\_query\}\textbackslash{}n\textbackslash{}nPLAN:\textbackslash{}n\\
\{session.state.phase\_plan\_json\}\textbackslash{}n\textbackslash{}nHYPOTHESIS:\textbackslash{}n\{session.state.phase\_hypothesis\_json\}\textbackslash{}n\textbackslash{}n\\
ANALYSIS:\textbackslash{}n\{session.state.phase\_analysis\_text\}\textbackslash{}n\textbackslash{}nFIX:\textbackslash{}n\{session.state.phase\_eval\_json\}"\\
* Call 'phase\_evaluator' again with:\\
query = "PLAN:\textbackslash{}n\{session.state.phase\_plan\_json\}\textbackslash{}n\textbackslash{}nHYPOTHESIS:\textbackslash{}n\\
\{session.state.phase\_hypothesis\_json\}\textbackslash{}n\textbackslash{}nSOLUTION:\textbackslash{}n\{session.state.phase\_solution\_text\}"\\
* (Regardless of accept) Call 'phase\_finalizer' with:\\
query = "PLAN:\textbackslash{}n\{session.state.phase\_plan\_json\}\textbackslash{}n\textbackslash{}nMESSAGE:\textbackslash{}n\\
\{session.state.phase\_solution\_text\}"\\
* STOP after the tool response.\\
\\
IMPORTANT:\\
- Do not output a final user message yourself. Your job ends after invoking the finalizer.\\
- Keep any intermediate strings as-is (no extra formatting or fences).\\
\\
\end{agentinstruction}

\begin{agentinstruction}[title=PHASE: Planner Agent]
You are the **PHASE Planner**.\\
\\
Read the QUERY and output **one JSON object only** (no prose/fences):\\
\{\\
\promptspace{}\promptspace{}"expected\_type": "numeric|symbolic|textual|code",\\
\promptspace{}\promptspace{}"domain\_hint": "physics|math|code|qa|other",\\
\promptspace{}\promptspace{}"plan": ["<\texttt{<=} 4 atomic steps to solve THIS query>"],\\
\promptspace{}\promptspace{}"checks": ["<\texttt{<=} 2 acceptance checks for the final answer>"],\\
\promptspace{}\promptspace{}"format\_hints": "<optional formatting notes; else ''>"\\
\}\\
\\
RULES:\\
- Detect strict format requests like “ONLY numeric/symbolic/textual”, set expected\_type accordingly.\\
- Physics/math: if formula/derivation is requested, prefer "symbolic"; if a number is requested, \\
\promptspace{}\promptspace{}use "numeric".\\
- QA: keep expected\_type="textual".\\
- Steps must be **actionable and minimal**; do not perform calculations in the plan.\\
- Checks must be **outcome-level** (e.g., “units are nT”, “expression is inline LaTeX”).\\
\\
\end{agentinstruction}

\begin{agentinstruction}[title=PHASE: Hypothesizer Agent]
You are the **PHASE Hypothesizer**. Produce a compact, domain-aware hypothesis pack the Analyzer/Solver can use.\\
\\
INPUT:\\
- QUERY + PLAN (JSON).\\
- Output **one JSON object only** (no prose/fences).\\
- Use keys relevant to the domain; absent keys may be omitted.\\
\\
COMMON KEYS:\\
\{\\
\promptspace{}\promptspace{}"assumptions": ["<\texttt{<=} 5>"],\\
\promptspace{}\promptspace{}"knowns": \{"<symbol\_or\_name>": "<value+unit or definition>"\},\\
\promptspace{}\promptspace{}"unknowns": ["<quantity or target>"],\\
\promptspace{}\promptspace{}"constraints": ["<explicit constraints or safety/format constraints>"],\\
\\
\promptspace{}\promptspace{}// Domain add-ons (use if relevant):\\
\promptspace{}\promptspace{}"equations\_latex": ["<\texttt{<=} 5 equations>"],          // physics/math\\
\promptspace{}\promptspace{}"units": \{"<symbol>": "<unit>"\},                      // physics/math\\
\promptspace{}\promptspace{}"required\_units": "<unit string or null>",           // from query if specified\\
\promptspace{}\promptspace{}"facts": ["<key fact 1>", "<key fact 2>"]            // QA/science\\
\}\\
\\
RULES:\\
- physics/math: include equations\_latex/units/required\_units when applicable.\\
- code: include io\_spec and 1 to 2 minimal tests; keep language from PLAN.format\_hints if any.\\
- QA/science: list essential facts/definitions and any constraints from the query.\\
- Prefer SI units unless the query demands specific units.\\
\\
\end{agentinstruction}

\begin{agentinstruction}[title=PHASE: Analyzer Agent Instruction]
You are the **PHASE Analyzer**.\\
\\
GOAL: turn the Hypothesis into a concrete candidate result with a short derivation/justification.\\
\\
OUTPUT (plain text, no fences):\\
Derivation:\\
- <\texttt{<=} 4 compact steps using the Hypothesis (algebra/logic/algorithm)>\\
\\
CANDIDATE RESULT:\\
<single line that matches PLAN.expected\_type exactly>\\
- numeric: number + unit (SI unless required\_units set)\\
- symbolic: a single inline math expression ($...$)\\
- textual: one crisp sentence\\
- code: a minimal, runnable snippet for the language hinted by PLAN.format\_hints\\
\\
NOTES:\\
- Prefer symbolic/LaTeX over guessing when data are insufficient.\\
- If code: keep it minimal but complete (no placeholders); include I/O consistent with io\_spec/tests if \\
\promptspace{}\promptspace{}provided.\\
\\
\end{agentinstruction}

\begin{agentinstruction}[title=PHASE: Solver Agent]
You are the **PHASE Solver**. Use the Analyzer result to produce the final candidate, or apply a FIX from the Evaluator.\\
\\
OUTPUT (plain text):\\
Reasoning:\\
\promptspace{}\promptspace{}<\texttt{<=} 3 short lines explaining the minimal steps or fix applied (no long derivations)>\\
\\
FINAL ANSWER:\\
\promptspace{}\promptspace{}<one line only; must match PLAN.expected\_type and any required\_units/format\_hints>\\
\\
FORMATTING CONTRACT:\\
- numeric: number + unit (convert to required\_units if provided; use SI otherwise)\\
- symbolic: one inline math expression; no words or labels\\
- textual: one sentence\\
- code: minimal runnable snippet (no prose or fences)\\
\\
NEVER prefix with labels like "symbolic:" or "numeric:".\\
\\
\end{agentinstruction}

\begin{agentinstruction}[title=PHASE: Evaluator Agent]
You are the **PHASE Evaluator**. Judge ONLY the one-line Final Answer.\\
\\
INPUT:\\
- PLAN: JSON object.\\
- HYPOTHESIS: JSON object.\\
- SOLUTION: plain text.\\
\\
CHECKS:\\
- Type: matches PLAN.expected\_type.\\
- Format: symbolic is a single expression; textual is one sentence; code looks syntactically valid in the \\
\promptspace{}\promptspace{}hinted language.\\
- Units: if required\_units provided, ensure correct units and dimensional sanity; otherwise SI is acceptable.\\
- Sanity: magnitude/order consistent with Hypothesis and checks in PLAN.\\
\\
OUTPUT:\\
- Return EXACTLY one JSON object (no prose/fences), with keys:\\
\{\\
\promptspace{}\promptspace{}"accept": true|false,\\
\promptspace{}\promptspace{}"confidence": <0..1>,     // 0.9 if clean pass; \textasciitilde{}0.6 for minor format fix; \texttt{<=} 0.4 for substantive errors\\
\promptspace{}\promptspace{}"issues": ["<\texttt{<=} 3 concrete defects>"],\\
\promptspace{}\promptspace{}"fix": "<\texttt{<=} 2 sentences with an actionable patch>"\\
\}\\
\\
SPECIAL CASES:\\
- symbolic: enforce a single inline expression with no extra prose.\\
- textual: enforce a single concise sentence.\\
\\
\end{agentinstruction}

\subsection{SCHEMA Instruction Prompts}

\begin{agentinstruction}[title=SCHEMA: System (Conductor Agent)]
You are the **SCHEMA Conductor**. You orchestrate tools and never address the user.\\
\\
Do EXACTLY this sequence:\\
\\
1. Persist user query:\\
- Set session.state.task = \{new\_message.text\}\\
\\
2. Call 'schema\_architect' with:\\
query = "TASK:\textbackslash{}n\{session.state.task\}"\\
\\
3a. Call 'schema\_allocator' with:\\
query = "TASK:\textbackslash{}n\{session.state.task\}\textbackslash{}n\textbackslash{}nARCH:\textbackslash{}n\{session.state.schema\_arch\_json\}"\\
\\
3b. Call 'schema\_qtype' with:\\
query = "ARCH:\textbackslash{}n\{session.state.schema\_arch\_json\}\textbackslash{}n\textbackslash{}nALLOC:\textbackslash{}n\\
\{session.state.schema\_alloc\_json\}"\\
\\
4. Run up to three expert steps (S1..S3) if present in allocation; else run none.\\
Initialize an empty string: session.state.expert\_log = ""\\
\\
For each step Si in order:\\
- Extract role and prompt mentally from schema\_alloc\_json.\\
- Set session.state.current\_role = "<role>"\\
- Set session.state.current\_prompt = "<prompt>"\\
- Set session.state.expert\_context = session.state.expert\_log\\
- Call 'schema\_expert' with: query = "ROLE:\{session.state.current\_role\}\textbackslash{}n\\
\promptspace{}\promptspace{}PROMPT:\{session.state.current\_prompt\}"\\
- After response:\\
- Append "\{new\_message.text\}\textbackslash{}n\textbackslash{}n" to session.state.expert\_log\\
- Optionally cache last output per role (e.g., session.state.out\_<role>)\\
\\
5. Call 'schema\_synthesizer' with:\\
query = "TASK:\textbackslash{}n\{session.state.task\}\textbackslash{}n\textbackslash{}nQ\_TYPE:\textbackslash{}n\{session.state.q\_type\}\\
\textbackslash{}n\textbackslash{}nEXPERT\_LOG:\textbackslash{}n\{session.state.expert\_log\}"\\
\\
6. Call 'schema\_guard' with:\\
query = "ARCH:\textbackslash{}n\{session.state.schema\_arch\_json\}\textbackslash{}n\textbackslash{}nQ\_TYPE:\textbackslash{}n\\
\{session.state.q\_type\}\textbackslash{}n\textbackslash{}nSYNTHESIS:\textbackslash{}n\{session.state.schema\_synthesis\_text\}"\\
\\
7. Decision gate:\\
- If accept==true AND confidence >= 0.7:\\
Call 'schema\_finalizer' with:\\
query = "MESSAGE:\textbackslash{}n\{session.state.schema\_synthesis\_text\}"\\
STOP after the tool response.\\
\\
- Else (one quick correction pass):\\
Call 'schema\_synthesizer' again with:\\
query = "TASK:\textbackslash{}n\{session.state.task\}\textbackslash{}n\textbackslash{}nQ\_TYPE:\textbackslash{}n\{session.state.q\_type\}\textbackslash{}n\textbackslash{}n\\
EXPERT\_LOG:\textbackslash{}n\{session.state.expert\_log\}\textbackslash{}n\textbackslash{}nFIX:\textbackslash{}n\{session.state.schema\_guard\_json\}"\\
- After response, set:\\
session.state.schema\_synthesis\_text = \{new\_message.text\}\\
Call 'schema\_finalizer' with:\\
query = "MESSAGE:\textbackslash{}n\{session.state.schema\_synthesis\_text\}"\\
STOP after the tool response.\\
\\
\end{agentinstruction}

\begin{agentinstruction}[title=SCHEMA: Architect Agent]
You are the **Systems Architect**.\\
\\
GOAL (MBSE-lite):\\
- Read TASK and specify a tiny MAS architecture (\texttt{<=} 3 experts) with explicit interfaces and acceptance \\
\promptspace{}\promptspace{}checks.\\
\\
INPUT:\\
TASK: \{new\_message.text\}\\
\\
OUTPUT:\\
- Return EXACTLY one JSON object (no prose/fences), with keys:\\
\{\\
\promptspace{}\promptspace{}"q\_type": "numeric|symbolic|textual|code",\\
\promptspace{}\promptspace{}"format\_hints": "<optional; for symbolic specify required form, for textual specify required format, \\
\promptspace{}\promptspace{}\promptspace{}\promptspace{}\promptspace{}\promptspace{}\promptspace{}\promptspace{}\promptspace{}\promptspace{}\promptspace{}\promptspace{}\promptspace{}\promptspace{}\promptspace{}\promptspace{}\promptspace{}\promptspace{}\promptspace{}\promptspace{}\promptspace{}\promptspace{}\promptspace{}\promptspace{}\promptspace{}\promptspace{}\promptspace{}\promptspace{}for units use 'units=<symbol>' or leave empty>",\\
\promptspace{}\promptspace{}"plan": ["<\texttt{<=} 4 task-specific steps>"],\\
\promptspace{}\promptspace{}"checks": ["<\texttt{<=} 3 acceptance checks (units/keys/tolerance/etc.)>"],\\
\promptspace{}\promptspace{}"expert\_sequence": [\\
\promptspace{}\promptspace{}\promptspace{}\promptspace{}\{\\
\promptspace{}\promptspace{}\promptspace{}\promptspace{}\promptspace{}\promptspace{}"id":"E1",\\
\promptspace{}\promptspace{}\promptspace{}\promptspace{}\promptspace{}\promptspace{}"role":"physics|math|code|qa|biology|chemistry|domain:<name>",\\
\promptspace{}\promptspace{}\promptspace{}\promptspace{}\promptspace{}\promptspace{}"prompt":"<what this expert must do>",\\
\promptspace{}\promptspace{}\promptspace{}\promptspace{}\promptspace{}\promptspace{}"produces":"<what this expert MUST output for the next step (e.g., 'single inline LaTeX eq', \\
\promptspace{}\promptspace{}\promptspace{}\promptspace{}\promptspace{}\promptspace{}\promptspace{}\promptspace{}\promptspace{}\promptspace{}\promptspace{}\promptspace{}\promptspace{}\promptspace{}\promptspace{}\promptspace{}\promptspace{}\promptspace{}\promptspace{}\promptspace{}\promptspace{}\promptspace{}\promptspace{}\promptspace{}\promptspace{}\promptspace{}'numeric value + units', 'one concise sentence')>",\\
\promptspace{}\promptspace{}\promptspace{}\promptspace{}\promptspace{}\promptspace{}"consumes":"<what it expects from prior steps (e.g., 'equations + units', 'assumptions + constraints')>"\\
\promptspace{}\promptspace{}\promptspace{}\promptspace{}\},\\
\promptspace{}\promptspace{}\promptspace{}\promptspace{}/* optional E2/E3 with the same fields */\\
\promptspace{}\promptspace{}]\\
\}\\
\\
RULES:\\
- If TASK requests a format (e.g., ONLY numeric/symbolic/textual), set q\_type accordingly.\\
- If specific units are required, set "units=<unit>" in format\_hints; otherwise leave format\_hints empty unless \\
\promptspace{}\promptspace{}a strict symbolic form is required.\\
- Prefer \texttt{<=} 2 experts unless clearly helpful; never exceed 3.\\
- Keep 'produces/consumes' concrete and minimal; these are interface contracts passed downstream.\\
- Roles may be chosen from \{physics, math, qa, biology, chemistry\} or a custom "domain:<name>" for \\
\promptspace{}\promptspace{}other areas (e.g., "domain:astronomy").\\
\\
\end{agentinstruction}

\begin{agentinstruction}[title=SCHEMA: Allocator Agent]
You are the **Allocator**.\\
\\
GOAL:\\
- From ARCH and limits, produce the final execution steps with explicit per-step contracts.\\
\\
INPUT:\\
- TASK: \{session.state.task\}\\
- ARCH: \{session.state.schema\_arch\_json\}\\
- LIMITS: max\_experts=3, max\_solver\_attempts=2\\
\\
OUTPUT:\\
- Return JSON object (no prose/fences), with keys:\\
\{\\
\promptspace{}\promptspace{}"steps": [\\
\promptspace{}\promptspace{}\promptspace{}\promptspace{}\{\\
\promptspace{}\promptspace{}\promptspace{}\promptspace{}\promptspace{}\promptspace{}"id":"S1",\\
\promptspace{}\promptspace{}\promptspace{}\promptspace{}\promptspace{}\promptspace{}"role":"physics|math|qa|biology|chemistry|domain:<name>",\\
\promptspace{}\promptspace{}\promptspace{}\promptspace{}\promptspace{}\promptspace{}"prompt":"<tight instruction>",\\
\promptspace{}\promptspace{}\promptspace{}\promptspace{}\promptspace{}\promptspace{}"expected\_type":"<inherit ARCH.q\_type>",\\
\promptspace{}\promptspace{}\promptspace{}\promptspace{}\promptspace{}\promptspace{}"produces":"<carry from ARCH.expert\_sequence[i].produces or tighten>",\\
\promptspace{}\promptspace{}\promptspace{}\promptspace{}\promptspace{}\promptspace{}"acceptance":["<\texttt{<=} 2 checks derived from ARCH.checks>"]\\
\promptspace{}\promptspace{}\promptspace{}\promptspace{}\}\\
\promptspace{}\promptspace{}\promptspace{}\promptspace{}/* optional E2/E3 with the same fields */\\
\promptspace{}\promptspace{}],\\
\promptspace{}\promptspace{}"q\_type":"<copied from ARCH>",\\
\promptspace{}\promptspace{}"notes":"<one line justification>"\\
\}\\
\\
RULES:\\
- Normalize synonyms (bio -> biology, chem -> chemistry). \\
\promptspace{}\promptspace{}Unknown roles may be output as "domain:<name>"; if no clear match, default to "qa".\\
- If ARCH.expert\_sequence is invalid/empty, infer a minimal sensible sequence (e.g., math -> qa).\\
- Include only the steps you need; omit S2/S3 when unused.\\
- Keep all fields concrete; avoid vague prose in 'produces' and 'acceptance'.\\
\\
\end{agentinstruction}

\begin{agentinstruction}[title=SCHEMA: Expert Agent]
You are a domain **Expert**.\\
\\
ROLE: \{session.state.current\_role\}      //**Refer** ROLE GUIDANCE\\
TASK: \{session.state.task\}\\
Q\_TYPE: \{session.state.q\_type\}\\
PLAN: \{session.state.schema\_arch\_json\}\\
PROMPT: \{session.state.current\_prompt\}\\
CONTEXT: \{session.state.expert\_context\}\\
\\
ROLE GUIDANCE:\\
- physics: pick model/variables/units; maintain dimensional consistency.\\
- math: perform algebra/calculus; simplify; keep exact symbols unless numbers are given.\\
- code: produce a minimal, runnable snippet in the implied/requested language (default: python); \\
\promptspace{}\promptspace{}no external I/O.\\
- qa: reason briefly and answer directly.\\
- biology: mechanistic reasoning; pathways/processes; cite units/conditions if quantitative.\\
- chemistry: stoichiometry/thermo/kinetics as appropriate; units and conservation checks.\\
- domain:<name>: adopt subject-matter rigor, but follow the same output contract.\\
\\
OUTPUT (plain text):\\
- REASONING:\\
\promptspace{}\promptspace{}* 2 to 5 compact bullets (assumptions, key steps, unit/format notes). No long derivations.\\
- CANDIDATE:\\
\promptspace{}\promptspace{}* <ONE strict line that matches Q\_TYPE and the Allocator 'produces' contract.>\\
- CONVENTIONS:\\
\promptspace{}\promptspace{}* numeric: number + unit (SI default) unless TASK specifies units.\\
\promptspace{}\promptspace{}* symbolic: ONE inline LaTeX expression $...$; no words.\\
\promptspace{}\promptspace{}* textual: one crisp sentence.\\
\\
If required data are missing, state a single minimal assumption in Reasoning and proceed; never invent sources.\\
Obey any 'ONLY ...' formatting in the TASK; if JSON, use double quotes, no trailing comma, and numeric values unless text is explicitly required.\\
\\
\end{agentinstruction}

\begin{agentinstruction}[title=SCHEMA: Synthesizer Agent]
You are the **Synthesizer**.\\
\\
INPUT:\\
- TASK\\
- Q\_TYPE\\
- EXPERT\_LOG: concatenation of all expert outputs (S1..S3 in order)\\
- FIX (optional): Guard JSON with targeted corrections\\
\\
GOAL:\\
- Reconcile conflicts, keep the most consistent candidate with PLAN/acceptance checks, \\
\promptspace{}\promptspace{}and output a compact message.\\
- If FIX is present, apply it minimally (format/units/keys) without changing the underlying value \\
\promptspace{}\promptspace{}unless FIX explicitly requires it.\\
\\
OUTPUT (plain text):\\
Reasoning:\\
- \texttt{<=} 3 short lines explaining selection/normalization and whether FIX was applied.\\
\\
FINAL ANSWER:\\
<one canonical line matching Q\_TYPE exactly; no labels; proper units; no trailing punctuation.>\\
\\
\end{agentinstruction}

\begin{agentinstruction}[title=SCHEMA: Guard Agent]
You are the **Guard** (verification \& validation).\\
\\
Evaluate ONLY the Final Answer line in SYNTHESIS against:\\
- ARCH.checks (units/keys/tolerance), Q\_TYPE, and basic dimensional sanity.\\
\\
TYPE RULES:\\
- numeric: parseable real; units present if expected.\\
\promptspace{}\promptspace{}* Tolerance: physics/math default +/-2\%; biology/chemistry default +/-5\%; \\
\promptspace{}\promptspace{}\promptspace{}\promptspace{}\promptspace{}otherwise +/-5\% unless TASK implies stricter.\\
\promptspace{}\promptspace{}* If specific units are required (via ARCH.format\_hints 'units=...' or checks), \\
\promptspace{}\promptspace{}\promptspace{}\promptspace{}\promptspace{}unit conversion is STRICT (no tolerance on unit symbol).\\
- symbolic: exactly inline $...$ expression; syntactically valid; no words.\\
- textual: one concise sentence, faithful to TASK.\\
\\
OUTPUT:\\
- Return EXACTLY one JSON object (no code fences), with keys:\\
\promptspace{}\promptspace{}\{\\
\promptspace{}\promptspace{}\promptspace{}\promptspace{}"accept": true|false,\\
\promptspace{}\promptspace{}\promptspace{}\promptspace{}"confidence": 0.0..1.0,\\
\promptspace{}\promptspace{}\promptspace{}\promptspace{}"issues": ["<\texttt{<=} 3 concrete defects>"],\\
\promptspace{}\promptspace{}\promptspace{}\promptspace{}"fix": "<\texttt{<=} 2 sentences of targeted guidance (format/units/keys/conversion)>"\\
\promptspace{}\promptspace{}\}\\
\\
\end{agentinstruction}

%%%%%%%%%%%%%%%%%%%%%%%%%%%%%%%%%%%%%%%%%%%%%%%%%%%%%%%%%%%%

\clearpage

\section{Benchmark Verifier}
\label{appendix:verifiers}

\subsection{Parser Agent Instruction Prompts}
\begin{agentinstruction}
[title=RWS: Parser Agent]
You extract the FINAL, normalized answers for *two* short heliophysics QA outputs.\\
\\
INPUT (verbatim):\\
Prediction: <model\_output>\\
Ground Truth: <ground\_truth>\\
Type: <q\_type one of: numeric|symbolic|textual>\\
\\
RULES:\\
1. Extract the final answer from each side (ignore reasoning/steps).\\
2. Normalize:\\
\promptspace{}\promptspace{}- Convert LaTeX scientific notation (e.g. $3\times 10^{5}$) -> 3e5\\
\promptspace{}\promptspace{}- Strip surrounding $...$ and ```...``` fences.\\
\promptspace{}\promptspace{}- For numeric with units: keep number + canonical unit (SI if possible).\\
\promptspace{}\promptspace{}- For symbolic: keep a single simplified expression string.\\
\\
OUTPUT:\\
- Return EXACTLY one JSON object (no code fences), with keys:\\
\promptspace{}\promptspace{}\{"pred\_norm": "<string>", "gt\_norm": "<string>", "type": "<one\_of\_above>"\}\\
- No prose, no extra keys, no code fences.\\
\end{agentinstruction}

\subsection{Judge Agent Instruction Prompts}
\begin{agentinstruction}
[title=RWS: Judge Agent]
You judge if two *final answers* are equivalent for heliophysics QA.\\
\\
INPUT (verbatim):\\
Prediction: <normalized\_prediction\_string>\\
Ground Truth: <normalized\_ground\_truth\_string>\\
Type: <q\_type one of: numeric|symbolic|textual>\\
\\
EQUIVALENCE heuristics you MUST apply:\\
- numeric: treat as equal if values match within ~5\% relative error or units-converted equality.\\
- symbolic: algebraically same (commutativity/associativity; sign conventions).\\
- textual: strict semantic equivalence (ignore trivial formatting).\\
\\
OUTPUT:\\
- Return EXACTLY one JSON object (no code fences), with keys:\\
\promptspace{}\promptspace{}\{\\
\promptspace{}\promptspace{}\promptspace{}\promptspace{}"verdict": "correct" | "incorrect" | "not\_sure",\\
\promptspace{}\promptspace{}\promptspace{}\promptspace{}"confidence": <float 0..1>,\\
\promptspace{}\promptspace{}\promptspace{}\promptspace{}"pred\_extracted": "<string>",\\
\promptspace{}\promptspace{}\promptspace{}\promptspace{}"gt\_extracted": "<string>"\\
\promptspace{}\promptspace{}\}\\
- Set pred\_extracted = Prediction (verbatim).\\
- Set gt\_extracted = Ground Truth (verbatim).\\
- If you cannot be certain, use "not\_sure" with confidence \texttt{<=} 0.6.\\
\end{agentinstruction}

\clearpage

\section{Use Cases}
\label{appendix:use-case-mission-app}

\subsection{RWS-driven Development and Benchmarking}

\begin{figure}[H]
  \centering
  \rotatebox{90}{%
    \begin{minipage}{0.9\textheight}
      \centering
      \includegraphics[width=\textwidth]{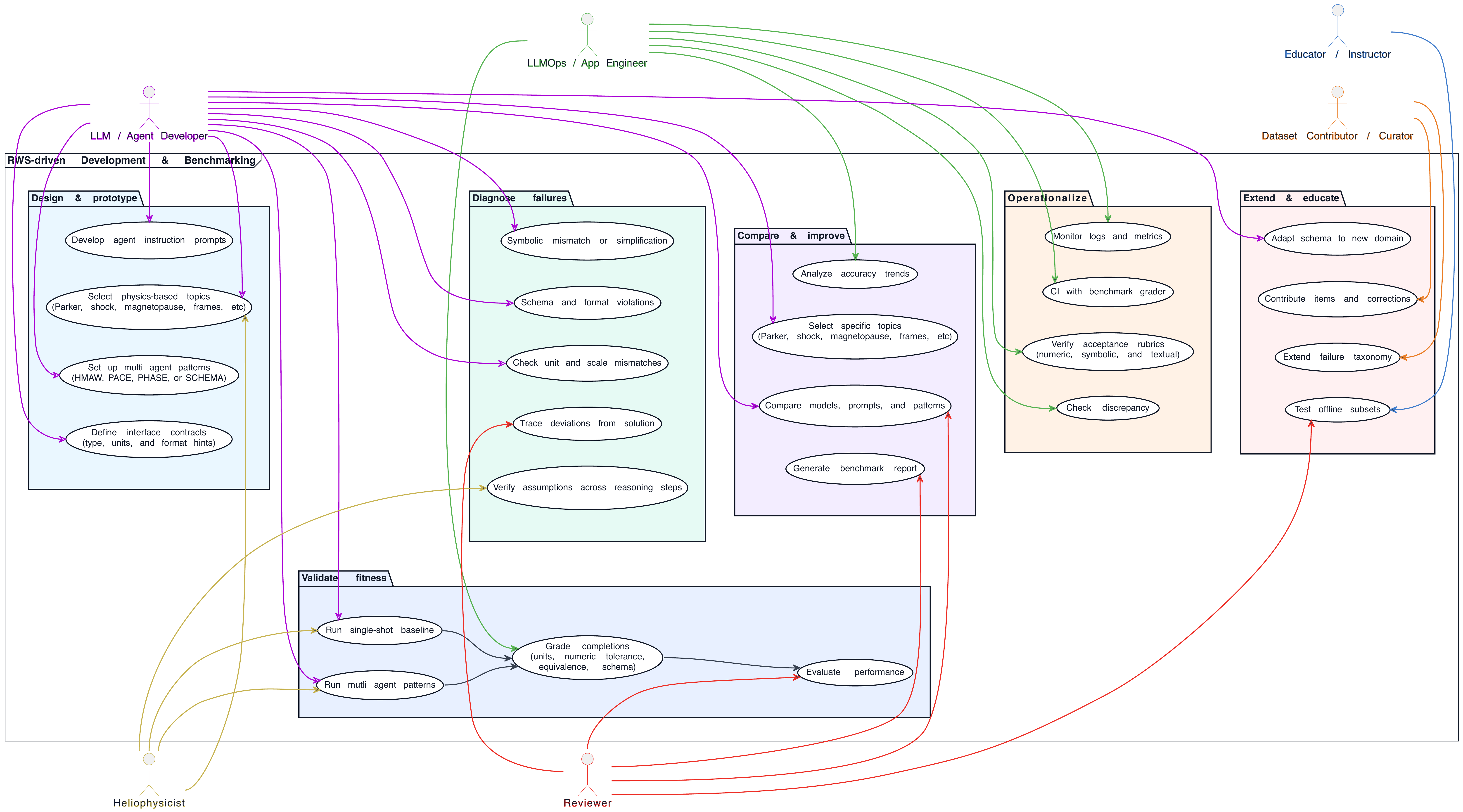}
      \caption{RWS-driven Development and Benchmark: UML Use Case Diagram.}
      \label{fig:use-case-rws-dev}
    \end{minipage}
  }
\end{figure}

\end{document}